\documentclass[aos,MSNbibl,seceqn,dvips]{arximspdf}
\usepackage[linesnumbered,ruled,vlined]{algorithm2e}
\usepackage{dcolumn}
\usepackage{graphicx}
\usepackage{breakurl}
\doi{10.1214/13-AOS1125} 
\volume{41}
\issue{1}
\pubyear{2013}
\firstpage{1742}
\lastpage{1779}

\makeatletter

\newcolumntype{d}[1]{D{.}{.}{#1}}

\newcommand{\rv}{\mbox{\textbf{\textit{rv}}}}

\newcommand{\eqref}[1]{(\ref{#1})}

\newcommand{\cal}{\mathcal}

\newproclaim{definition}{Definition}

\newtheorem{theorem}{Theorem}
\newtheorem{lemma}{Lemma}
\newtheorem{proposition}{Proposition}
\newtheorem{corollary}{Corollary}

\makeatother

\begin{document}
\begin{frontmatter}

\title{Reversible MCMC on Markov equivalence classes of sparse
directed acyclic graphs\thanksref{T11}}
\runtitle{Reversible MCMC on MECs of sparse DAGs}

\thankstext{T11}{Supported in part by NSFC (11101008, 11101005, 71271211),
973 Program-2007CB814905, DPHEC-20110001120113, US NSF Grants DMS-11-07000,
DMS-09-07632, DMS-06-05165, DMS-12-28246 3424, SES-0835531 (CDI), US ARO
grant W911NF-11-1-0114 and the Center for Science of Information
(CSoI), a US NSF Science and Technology Center, under Grant agreement
CCF-0939370.
This research was also supported by School of Mathematical Science,
the Center of Statistical Sciences, the Key Lab of Mathematical
Economics and Quantitative Finance (Ministry of Education) , the Key
lab of Mathematics and Applied Mathematics (Ministry od Education),
and the Microsoft Joint Lab on Statistics and information technology at
Peking University.}

\begin{aug}
\author[A]{\fnms{Yangbo} \snm{He}\corref{}\ead[label=e1]{heyb@math.pku.edu.cn}},
\author[A]{\fnms{Jinzhu} \snm{Jia}\ead[label=e2]{jzjia@math.pku.edu.cn}}
\and
\author[B]{\fnms{Bin} \snm{Yu}\ead[label=e3]{binyu@stat.berkeley.edu}}
\runauthor{Y. He, J. Jia and B. Yu}
\affiliation{Peking University, Peking University and
University of California, Berkeley}
\address[A]{Y. He\\
J. Jia\\
School of Mathematical Sciences\\
and Center of Statistical Sciences\\
Peking University\\
Beijing 100871\\
China \\
\printead{e1}\\
\hphantom{E-mail: }\printead*{e2}}
\address[B]{B. Yu\\
Department of Statistics \\
University of California\\
Berkeley, California 94720\\
USA\\
\printead{e3}} 
\end{aug}

\received{\smonth{9} \syear{2012}}
\revised{\smonth{4} \syear{2013}}

%
\begin{abstract}
Graphical models are popular statistical tools which are used to
represent dependent or causal complex systems. Statistically
equivalent causal or directed graphical models are said to belong to a
Markov equivalent class. It is of great interest to describe and
understand the space of such classes. However, with currently known
algorithms, sampling over such classes is only feasible for graphs with
fewer than approximately 20 vertices. In this paper, we design
reversible irreducible Markov chains on the space of Markov equivalent
classes by proposing a \textit{perfect} set of operators that determine
the transitions of the Markov chain. The stationary distribution of
a proposed Markov chain has a closed form and can be computed easily.
Specifically, we construct a concrete perfect set of operators on
sparse Markov equivalence classes by introducing appropriate conditions
on each possible operator. Algorithms and their accelerated versions
are provided to efficiently generate Markov chains and to explore
properties of Markov equivalence classes of sparse directed acyclic
graphs (DAGs) with thousands of vertices. We find experimentally that
in most Markov equivalence classes of sparse DAGs, (1)~most edges are
directed, (2) most undirected subgraphs are small and (3) the number
of these undirected subgraphs grows approximately linearly with the
number of vertices.
\end{abstract}

%

\end{frontmatter}

\section{Introduction}
Graphical models based on directed acyclic graphs\break (DAGs, denoted as
$\cal D$) are widely used to represent causal or dependent
relationships in various scientific investigations, such as
bioinformatics, epidemiology,
sociology and business \cite
{finegold2011robust,Friedman,heckerman1999bayesian,jansen2003bayesian,maathuis2009estimating,pearl2000causality,spirtes2001causation}.
A DAG encodes the independence and conditional
independence restrictions of variables. However, because different DAGs
can encode the same set of independencies or conditional
independencies, most of the time we cannot distinguish DAGs via
observational data~\cite{pearl1988probabilistic}.
A Markov equivalence class is used to represent all DAGs that encode
the same dependencies and independencies
\cite{andersson1997characterization,chickering2002learning,pearl1991theory}.
A Markov equivalence class can be visualized (or modeled) and uniquely
represented by a completed partial directed acyclic graph (completed
PDAG for short)~\cite{chickering2002learning}
which possibly contains both directed edges and undirected edges~\cite
{lauritzen2002chain}.
There exists a one-to-one correspondence between completed PDAGs and
Markov equivalence classes~\cite{andersson1997characterization}.
The completed PDAGs are also called essential graphs by Andersson et
al.~\cite{andersson1997characterization} and maximally oriented graphs
by Meek~\cite{meek1995causal}.

A \emph{set} of completed PDAGs can be used as a model space. The
modeling task is to discover a proper Markov equivalence class in the
model space \cite
{castelo2004learning,chickering1995learning,CY1999,dash1999hybrid,heckerman1995learning,madigan1996bayesian}.
Understanding the set of Markov equivalence classes is important and
useful for statistical causal modeling \cite
{gillispie2006formulas,gillispie2002size,Kalisch}.
For example, if the number of DAGs is large for Markov equivalence
classes in the model space,
searching
based on unique completed PDAGs could be substantially more efficient
than searching based on DAGs \cite
{chickering2002learning,munteanu2001eq,madigan1996bayesian}.
Moreover, if most completed PDAGs in the model space have many
undirected edges (with nonidentifiable directions), many interventions
might be needed to identify the causal directions \cite
{eberhardt2007interventions,he2008active}.

Because the number of Markov
equivalence classes increases superexponentially with the number
of vertices (e.g., more than $10^{18}$ classes with 10 vertices) \cite
{gillispie2002size}, it is hard to study sets of Markov
equivalence classes. To our knowledge, only completed PDAGs with a
small given number of vertices ($\leq$10) have been studied thoroughly
in the literature \cite
{gillispie2006formulas,gillispie2002size,pena2007approximate}.
Moreover, these studies focus on the size of Markov equivalence classes,
which is defined as the number of DAGs in a Markov equivalence class.
Gillispie and Perlman~\cite{gillispie2002size} obtain the true size
distribution of all Markov equivalence classes with a given number (10
or fewer) of vertices by listing all classes. Pe\~na \cite
{pena2007approximate} designs a Markov chain to estimate the proportion
of the equivalence classes containing only one DAG for graphs with 20
or fewer vertices.

In recent years, sparse graphical models have become popular tools for
fitting high-dimensional multivariate data. The sparsity assumption
introduces restrictions on the model space; a standard restriction is
that the number of edges in the graph be less than a small multiple of
the number of vertices. It is thus both interesting and important to be
able to explore the properties of subsets of graphical models,
especially with sparsity constraints on the edges.\looseness=-1

In this paper, we propose a reversible irreducible Markov chain on Markov
equivalence classes.
We first introduce a perfect set of operators that determine the
transitions of the chain. Then we obtain the stationary distribution of
the chain by counting (or estimating) all possible transitions for each
state of the chain. Finally, based on the stationary distribution of the
chain (or estimated stationary distribution), we re-weigh the samples
from the chain.
Hence these reweighed samples can be seen as uniformly (or
approximately uniformly) generated from
the Markov equivalence classes of interest. Our proposal allows the
study of properties of the sets that contain sparse Markov equivalence
classes in a computationally efficient manner for sparse graphs with
thousands of vertices.\vspace*{-2pt}

\subsection{A Markov equivalence class and its representation}
\label{sub11}

In this section, we give a short overview for the representations of a
Markov equivalence class.

A graph ${\cal G}$ is defined as a pair $(V,
E)$, where $V=\{x_{1},\ldots,x_{p}\}$ denotes the
vertex set with $p$ variables, and $E$ denotes the edge set. Let
$n_{\cal G}=|E|$ be the number of edges in $\cal G$. A directed
(undirected) edge is denoted as $\rightarrow
$ or $\leftarrow$ ($-$).
A graph is directed (undirected) if all of its edges are directed (undirected).
A sequence $(x_{1},x_{2},\ldots,x_{k})$ of distinct vertices is called
a \textit{path}
from $x_{1}$ to $x_{k}$ if either $x_{i} \rightarrow
x_{i+1}$ or $x_{i} - x_{i+1}$ is in $E$ for all $i=1,\ldots,
k-1$. A path is partially directed if at least one edge in it is
directed. A path is directed (undirected) if all edges are directed
(undirected).
A \textit{cycle} is a
path from a vertex to itself.

A \textit{directed acyclic graph} (DAG), denoted by $\cal D$, is a
directed graph which does not
contain any directed cycle.
Let $\tau$ be a subset of
$V$. The \textit{subgraph} ${\cal D}_{\tau
}=(\tau,E_{\tau})$ induced by the subset $\tau$ has
vertex set $\tau$ and edge set
$E_{\tau}$, the subset of $E$ which contains the
edges with both vertices in $\tau$.
A subgraph $x\rightarrow z\leftarrow y$ is called a \textit
{$v$-structure} if there is no edge between $x$ and $y$. A~\textit
{partially directed acyclic graph} (PDAG), denoted by $\cal{P}$, is a
graph with no directed cycle.

A graphical model consists of a DAG and a joint probability distribution.
With the graphical model, in general, the conditional independencies implied
by the joint probability distribution can be read from the DAG. 
\textit{A~Markov
equivalence class} (MEC) is a set of DAGs that encode the same set of
independencies or conditional independencies. Let the \textit
{skeleton} of an arbitrary graph $\cal G$ be the undirected graph with
the same vertices and edges as $\cal G$, regardless of their
directions. Verma and Pearl~\cite{verma1990equivalence} proved the
following characterization of Markov equivalence classes:\vspace*{-2pt}
%
\begin{lemma}[(Verma and Pearl~\cite{verma1990equivalence})]\label{makovverma}
Two DAGs are Markov equivalent if and only if they have
the same
skeleton and the same $v$-structures.
\end{lemma}
%

This lemma implies that, among DAGs in an equivalence class, some edge
orientations may vary, while others will be preserved (e.g.,\vadjust{\goodbreak} those
involved in a $v$-structure). Consequently, a Markov equivalence class
can be represented uniquely by a \textit{completed PDAG}, defined as
follows:

\begin{definition}[(Completed PDAG~\cite{chickering2002learning})]
\label{defpdag}
The \textit{completed PDAG} of a\break DAG~${\cal D}$, denoted as ${\cal C}$,
is a PDAG that has the same skeleton as ${\cal D}$, and an edge
is directed in ${\cal C}$ if and only if it has the same orientation in
every equivalent DAG of $\cal D$.
\end{definition}
According to Definition~\ref{defpdag} and Lemma~\ref{makovverma}, a
completed PDAG of a DAG ${\cal D}$ has the same skeleton as ${\cal D}$,
and it keeps at least the directed edges that occur in the $v$-structures
of ${\cal D}$.
Another popular name of a completed PDAG is ``essential graph''
introduced by Andersson et al.~\cite{andersson1997characterization},
who introduce four necessary and sufficient conditions for a graph to
be an essential graph; see them in Lemma~\ref{essential}, Appendix~\ref{appA.1}.
One of the conditions shows that all directed edges in a completed PDAG
must be ``strongly protected,'' defined as follows:
%
\begin{definition}
\label{spro}
Let ${\cal G}=(V,E)$ be a graph. A directed edge $v\to u\in E$ is
strongly protected in $\cal G$ if $v\to u\in E$ occurs in at least one
of the four induced subgraphs of $\cal G$ in Figure~\ref{spro111}.

\begin{figure}

\includegraphics{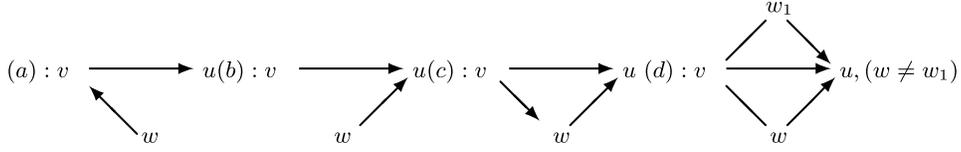}

\caption{Four configurations where $v\to u$ is strongly protected in
$\cal G$.} \label{spro111}
\end{figure}
\end{definition}
%

If we delete all directed edges from a completed PDAG, we are left
with several isolated undirected subgraphs. Each isolated undirected
subgraph is a \textit{chain component} of the completed PDAG.
Observational data is not sufficient to learn the directions of
undirected edges of a completed PDAG; one must perform additional
intervention experiments. In general, the size of a chain component is
a measure of ``complexity'' of causal learning; the larger the chain
components are, the more interventions will be necessary to learn the
underlying causal graph~\cite{he2008active}.

In learning graphical models~\cite{chickering2002learning} or studying
Markov equivalence classes~\cite{pena2007approximate}, Markov chains on
completed PDAGs play an important role. We briefly introduce the
existing methods to construct Markov chains on completed PDAGs in the
next subsection.

\subsection{Markov chains on completed PDAGs}
\label{subs22}

To construct a Markov chain on completed PDAGs, we need to generate the
transitions among them. In general, an \textit{operator} that can
modify the initial completed\vadjust{\goodbreak} PDAG locally can be used to carry out a
transition \cite
{perlman2001graphical,chickering2002learning,munteanu2001eq,pena2007approximate}.
Let $\cal C$ be a completed PDAG. We consider six types of operators
on $\cal C$: inserting an undirected edge (denoted by InsertU ),
deleting an undirected edge (DeleteU), inserting a directed edge
(InsertD), deleting a directed edge (DeleteD), making a $v$-structure
(MakeV) and removing a $v$-structure (\mbox{RemoveV}). We call InsertU, DeleteU,
InsertD, DeleteD, MakeV and RemoveV the \textit{types} of operators. An
operator on a given completed PDAG is determined by two parts: its type
and the modified edges. For example, the operator ``InsertU $x-y$'' on
$\cal C$ represents inserting an undirected edge $x-y$ to $\cal C$, and
$x-y$ is the modified edge of the operator.
A \textit{modified graph} of an operator is the same as the initial
completed PDAG, except for the modified edges of the operator.
A~modified graph might (not) be a completed PDAG; see Example~1 in
Section~2.1, of the Supplementary Material~\cite{hesupp}.

Madigan et al.~\cite{madigan1996bayesian}, Perlman \cite
{perlman2001graphical} and Pe\~na~\cite{pena2007approximate} introduce
several Markov chains based on the modified graphs of operators. At
each state of these Markov chains, say $\cal C$, they move to the
modified graph of an operator on $\cal C$ only when the modified graph
happens to be a completed PDAG, otherwise, stay at $\cal C$. In order
to move to new completed PDAGs, Madigan et~al. \cite
{madigan1996bayesian} search the operators whose modified graphs are
completed PDAG by checking Andersson's conditions \cite
{andersson1997characterization} one by one. Perlman \cite
{perlman2001graphical} introduces an alternative search approach that
is more efficient by ``exploiting further'' Andersson's conditions.

When the modified graph of an operator on $\cal C$ is not a completed
PDAG, the operator might result in a transition from one completed PDAG
$\cal C$ to another. This operator also results in a ``valid''
transition. To obtain valid transitions, Chickering \cite
{chickering2002learning,chickering2003optimal} introduces the concept
of \textit{validity} for an operator on $\cal C$. Before defining
``valid operator,'' we need a concept \textit{consistent extension}. A
\textit{consistent extension} of a PDAG $\cal P$ is a directed acyclic
graph (DAG) on the same underlying set of edges, with the same
orientations on the directed edges of $\cal P$ and the same set of
\mbox{$v$-structures}~\cite{dor1992simple,verma1992algorithm}. According to
Lemma~\ref{makovverma}, all consistent extensions of a PDAG~$\cal P$,
if they exist, belong to a unique Markov equivalence class. Hence if
the modified graph of an operator is a PDAG and has a consistent
extension, it can result in a completed PDAG that corresponds to a unique
Markov equivalence class. We call it the \textit{resulting} completed
PDAG of the operator. Now a valid operator is defined as below.

\begin{definition}[(Valid operator)]\label{defvalid}
An operator on $\cal C$ is \textit{valid} if (1) the modified graph of
the operator is a PDAG and has a consistent extension, and (2) all
modified edges in the modified graph occur in the resulting completed
PDAG of the operator.
\end{definition}

The first condition in Definition~\ref{defvalid} guarantees that a
valid operator results in a completed PDAG.\vadjust{\goodbreak} The second condition
guarantees that the valid operator is ``effective;'' that is, the change
brought about by the operator occurs in the resulting completed PDAG.
Here we notice that the second condition is implied by the context in
Chickering~\cite{chickering2002learning}. Below we briefly introduce
how to obtain the resulting completed PDAG of a valid operator from the
modified graph.

Verma and Pearl~\cite{verma1992algorithm} and Meek \cite
{meek1995causal} introduce an algorithm for finding the completed PDAG
from a ``pattern'' (given skeleton and $v$-structures). This method can be
used to create the completed PDAG from a DAG or a PDAG. They first
undirect every edge, except for those
edges that participate in a $v$-structure. Then they choose one of the
undirected edges and direct it if the corresponding directed edge is
strongly protected, as shown in Figure~\ref{spro111}(a), (c) or (d).
The algorithm terminates when there is no undirected edge that can be directed.

Chickering~\cite{chickering2002learning} proposes an alternative
approach to obtain the completed PDAG of a valid operator from its
modified graph; see Example 2, Section~2.1 of the Supplementary
Material~\cite{hesupp}. The method includes two steps. The first step
generates a consistent extension (a DAG) of the modified graph (a PDAG)
using the algorithm described in Dor and Tarsi~\cite{dor1992simple}.
The second step creates a completed PDAG corresponding to the
consistent extension \cite
{chickering1995transformational,chickering2002learning}.
We describe Dor and Tarsi's algorithm and Chickering's algorithms in
Section 1 of the Supplementary Material~\cite{hesupp}.

The approach proposed by Chickering \cite
{chickering1995transformational,chickering2002learning} is ``more
complicated but more efficient''~\cite{meek1995causal} than Meek's
method described above. Hence when constructing a Markov chain, we use
Chickering's approach to obtain the resulting completed PDAG of a given
valid operator from its modified graph.

With a set of valid operators, a Markov chain on completed PDAGs can
be constructed. Let ${\cal S}_p$ be the set of all completed PDAGs with
$p$ vertices, $\cal S$ be a given subset of ${\cal S}_p$. For any
completed PDAG ${\cal C} \in{\cal S}$, let ${\cal O}_{\cal C}$ be a
set of valid operators of interest to be defined later on $\cal C$ in
equation \eqref{constraintedC}. A~set of
valid operators on $\cal S$ is defined as
%
\begin{equation}
\label{operatorset1} {\cal O}=\bigcup_{{\cal C}\in{\cal S}}{\cal
O}_{\cal C}.
\end{equation}
Here we notice that each operator in ${\cal O}$ is specific to the
completed PDAG that the operator applies to. A Markov chain $\{e_t\}$
on ${\cal S}$ based on the set $\cal O$ can be defined as follows.

\begin{definition}[(A Markov chain $\{e_t\}$ on ${\cal S}$)]\label{markovchain}
The Markov chain $\{e_t\}$ determined by a set of valid operators $\cal
O$ is generated as follows: start at an arbitrary completed PDAG,
denoted as $e_0={\cal C}_0\in{\cal S}$, and repeat the following steps
for $t=0,1,\ldots\,$:
\begin{longlist}[(2)]
\item[(1)]At the $t$th step we are at a completed PDAG $e_t$.
\item[(2)] We choose an operator $o_{e_t}$ uniformly from ${\cal
O}_{e_t}$; if the resulting completed PDAG ${\cal C}_{t+1}$ of
$o_{e_t}$ is in ${\cal S}$, move to ${\cal C}_{t+1}$ and set
$e_{t+1}={\cal C}_{t+1}$; otherwise we stay at $e_t$ and set $e_{t+1}={e_t}$.
\end{longlist}
\end{definition}


Given the same operator set, the Markov chain in Definition \ref
{markovchain} has more new transition states for any completed PDAG
than those based on the modified graphs of operators \cite
{madigan1996bayesian,perlman2001graphical,pena2007approximate}. This
is because some valid operators will result in new completed PDAGs even
if their modified graphs are not completed PDAGs. Consequently, the
transitions, which are generated by these operators, are not contained
in Markov chains based on the modified graphs.

The set $\cal S$ is the finite state space of chain $\{e_t\}$. Clearly,
the sequence of completed PDAGs $\{e_t\dvtx t=0,1,\ldots\}$ in Definition
\ref{markovchain} is a discrete-time Markov chain \cite
{lovasz1993random,norris1997markov}.
Let $p_{_{ {\cal C}{\cal C}'}}$ be the one-step transition probability
of $\{e_t\}$ from ${\cal C}$ to ${\cal C}'$ for any two completed PDAGs
${\cal C}$ and ${\cal C}'$ in~$\cal S$. A Markov chain $\{e_t\}$ is
\textit{irreducible} if it can reach any completed PDAG starting at any
state in~$\cal S$.
If $\{e_t\}$ is irreducible, there exists a unique distribution $\pi
=(\pi_{_{\cal C}}, {\cal C}\in{\cal S})$ satisfying balance equations
(see Theorems 1.7.7 and 1.5.6 in~\cite{norris1997markov})
%
\begin{equation}
\label{stationarydistribution} \pi_{_{{\cal C}}} =\sum
_{{\cal C}' \in{\cal S}}\pi_{_{{\cal
C}'}}p_{_{{\cal C}'{\cal C}}} \qquad\mbox{for all } {\cal
C} \in{\cal S}.
\end{equation}

An irreducible chain $e_t$ is \textit{reversible} if there exists a
probability distribution $\pi$ such that
%
\begin{equation}
\label{reversible} \pi_{_{\cal C}} p_{_{{\cal C}{\cal C}'}}=\pi
_{_{{\cal C}'}}
p_{_{{\cal
C}'{\cal C}}}\qquad \mbox{for all } {\cal C},{\cal C}' \in{\cal S}.
\end{equation}

It is well known that $\pi$ is the unique stationary distribution of
the discrete-time Markov chain $\{e_t\}$ if it is finite, reversible,
and irreducible; see Lem\-ma~1.9.2 in~\cite{norris1997markov}. Moreover,
the stationary probabilities $\pi_{_{{\cal C}}}$ can be calculated
efficiently if the Markov chain satisfies equation \eqref{reversible}.

The properties of the Markov chain $\{e_t\}$ given in Definition \ref
{markovchain}
depend on the operator set $\cal O$. To implement score-based searching
in the whole set of Markov equivalence classes, Chickering \cite
{chickering2002learning} introduces a set of operators with types of
InsertU, DeleteU, InsertD, DeleteD, MakeV or ReverseD (reversing the
direction of a directed edge), subject to some validity conditions.
Unfortunately, the Markov chain in Definition~\ref{markovchain} is not
reversible if the set of Chickering's operators is used. Our goal is to
design a reversible Markov chain, as it makes it easier to compute the
stationary distribution, and thereby to study the properties of a
subset of Markov equivalence classes.

In Section~\ref{randomwalk}, we first discuss the properties of an
operator set $\cal O$ needed to guarantee that the Markov chain is
reversible. Section~\ref{randomwalk} also explains how to use the
samples from the Markov chain to study properties of any given subset
of Markov equivalence classes. In Section~\ref{sparseMC} we focus on
studying sets of sparse Markov equivalence classes. Finally, in
Section~\ref{experiments}, we report the properties of directed edges
and chain components in sparse Markov equivalence classes with up to
one thousand of vertices.

\section{Reversible Markov chains on Markov equivalence classes}
\label{randomwalk}

Let $\cal S$ be any subset of the set ${\cal S}_p$
that contains all completed PDAGs with $p$ vertices, and $\cal O$ be a
set of operators on $\cal S$ defined in equation \eqref{operatorset1}.
As in Definition~\ref{markovchain}, we can obtain a Markov chain
denoted by $\{e_t\}$.
We first discuss four properties of $\cal O$ that guarantee that $\{
e_t\}$ is reversible and irreducible. They are \textit{validity},
\textit{distinguishability}, \textit{irreducibility} and \textit{reversibility}. We call
a set of operators \textit{perfect} if it satisfies these four
properties. Then we give the stationary distribution of $\{e_t\}$ when
$\cal O$ is perfect and show how to use $\{e_t\}$ to study properties
of $\cal S$.

\subsection{A reversible Markov chain based on a perfect set of operators}

Let $p_{_{ {\cal C}{\cal C}'}}$ be a one-step transition probability
of $\{e_t\}$ from ${\cal C}$ to ${\cal C}'$ for any two completed PDAGs
${\cal C}$ and ${\cal C}'$ in $\cal S$.
In order to formulate $p_{_{ {\cal C}{\cal C}'}}$ clearly, we
introduce two properties of $\cal O$: Validity and Distinguishability.

\begin{definition}[(Validity)]\label{validdef}
Given $\cal S$ and any completed PDAG $\cal C$ in ${\cal S}$, a set of
operators $\cal O$ on ${\cal S} $ is valid if for any operator
$o_{_{\cal C}}$ ($o$ without confusion below) in ${\cal O}_{\cal C}$,
$o$ is valid according to Definition~\ref{defvalid} and the resulting
completed PDAG obtained by applying $o$ to $\cal C$,
which is different from $\cal C$, is also in $\cal S$.
\end{definition}

According to Definition~\ref{validdef}, if a set of operators $\cal O$
on ${\cal S}$ is valid, we can move to a new completed PDAG in each
step of $\{e_t\}$ and the one-step transition probability of any
completed PDAG to itself is zero:
%
\begin{equation}
p_{_{{\cal C}{\cal C}}}=0\qquad \mbox{for any completed PDAG } {\cal C}\in
{\cal S}.
\end{equation}
For a set of valid operators $\cal O$ and any completed PDAG $\cal C$
in $\cal S$, we define the resulting completed PDAGs of the operators
in ${\cal O}_{\cal C}$ as the \textit{direct successors} of $\cal C$.
For any direct successor of $\cal C$, denoted by $\cal C'$, we obtain
$p_{_{ {\cal C}{\cal C}'}}$ clearly as in equation \eqref{pcc} if
$\cal O$ has the following property.

\begin{definition}[(Distinguishability)] \label{distdef}
A set of valid operators $\cal O$ on ${\cal S}$ is distinguishable if
for any completed PDAG $\cal C$ in ${\cal S}$, different operators in
${\cal O }_{\cal C}$ will result in different completed PDAGs.
\end{definition}

If $\cal O$ is distinguishable, for any direct successor of $\cal C$,
denoted by $\cal C'$, there is a unique operator in ${\cal O}_{\cal C}$
that can transform ${\cal C}$ to $\cal C'$. Thus, the number of
operators in ${\cal O}_{\cal C}$ is the same as the number of direct
successors of $\cal C$. Sampling operators from ${\cal O}_{\cal C}$
uniformly generates a uniformly random transition from $\cal C$ to its
direct successors. By denoting $M({\cal O}_{\cal C})$ as the number of
operators in ${\cal O}_{\cal C}$, we have
%
\begin{equation}
\label{pcc} p_{_{{\cal C}{\cal C}'}}=\cases{1/M({\cal
O}_{\cal C}), &\quad ${\cal C}'$ is a direct successor of ${\cal
C} \in{\cal S} $;
\cr
0, &\quad otherwise.}
\end{equation}

We introduce this property because it makes computation of { $p_{\cal
{CC}'}$} efficient: if $\cal O$ is distinguishable, we know $
p_{_{{\cal C}{\cal C}'}}$ right away from $M({\cal O}_{\cal
C})$.\vadjust{\goodbreak}

In order to make sure the Markov chain $\{e_t\}$ is irreducible and reversible,
we introduce two more properties of $\cal O$: irreducibility and reversibility.

\begin{definition}[(Irreducibility)] \label{irreducibility}
A set of operators $\cal O$ on ${\cal{S}}$ is irreducible if for any
two completed PDAGs ${\cal C},{\cal C}' \in{\cal{S}}$, there exists a
sequence of operators in ${\cal O}$ such that we can obtain ${\cal C}'$
from ${\cal C}$ by applying these operators sequentially.
\end{definition}
If $\cal O$ is irreducible, starting at any completed PDAG in ${\cal
{S}}$, we have positive probability to reach any other completed PDAG
via a sequence of operators in $\cal O$. Thus, the Markov chain $\{
{e_t}\}$ is irreducible.

\begin{definition}[(Reversibility)]\label{revdef}
A set of operators $\cal O$ on ${\cal{S}}$ is reversible if for any
completed PDAG ${\cal C}\in{\cal{S}}$ and any operator $o\in{\cal
O}_{{\cal C}}$ with ${\cal C}'$ being the resulting completed PDAG of
$o$, there is an operator $o' \in{\cal O}_{{\cal C}'}$ such that
${\cal C}$ is the resulting completed PDAG of $o'$.
\end{definition}

If the set of operators $\cal O$ on ${\cal{S}}$ is valid,
distinguishable and reversible, for any pair of completed PDAGs ${\cal
C},{\cal C}' \in{\cal S}$, ${\cal C}$ is also a direct successor of
${\cal C}'$ if ${\cal C}'$ is a direct successor of ${\cal C}$.
For any ${\cal C} \in{\cal S} $ and any of its direct successors~${\cal C}'$, we have
%
\begin{equation}
\label{onesteptran} p_{_{{\cal C}{\cal C}'}}=1/M({\cal O}_{\cal C})
\quad\mbox{and}\quad
p_{_{{\cal C}'{\cal C}}}=1/M({\cal O}_{{\cal C}'}).
\end{equation}

Let ${\cal T}=\sum_{{\cal C}\in{\cal S} }M({\cal O}_{\cal C})$, and
define a probability distribution as
%
\begin{equation}
\label{stationarydist} \pi_{_{\cal C}}=M({\cal O}_{\cal
C})/{\cal T}.
\end{equation}
Clearly, equation \eqref{reversible} holds for $\pi_{_{\cal C}}$ in
equation \eqref{stationarydist} if $\cal O$ is valid, distinguishable
and reversible. $\pi_{_{\cal C}}$ is the unique stationary distribution
of $\{e_t\}$ if it is also irreducible \cite
{AldoFill99,lovasz1993random,norris1997markov}.

In the following proposition, we summarize our results about the Markov
chain $\{e_t\}$ on ${\cal S} $, and give its stationary distribution.
%
\begin{proposition}[(Stationary distribution of $\{e_t\}$)]
\label{stationaryd}
Let $\cal S$ be any given set of completed PDAGs. The set of operators
is defined as ${\cal O}=\bigcup_{{\cal C}\in{\cal S}}{\cal O}_{\cal
C}$ where ${\cal O}_{\cal C}$ is a set of operators on ${\cal C}$ for
any $\cal C$ in ${\cal S}$. Let $M({\cal O}_{\cal C})$ be the number of
operators in ${\cal O}_{\cal C}$.
For the Markov chain $\{e_t\}$ on ${\cal S}$ generated according to
Definition~\ref{markovchain}, if
$\cal O$ is perfect, that is, the properties---validity,
distinguishability, reversibility and irreducibility---hold for ${\cal
O}$, then:
\begin{longlist}[(2)]
\item[(1)] the Markov chain $\{e_t\}$ is irreducible and reversible;
\item[(2)] the distribution $\pi_{_{\cal C}}$ in equation \eqref
{stationarydist} is the unique stationary distribution of $\{e_t\}$ and
$\pi_{_{\cal C}}\propto M({\cal O}_{\cal C})$.
\end{longlist}
\end{proposition}

The challenge is to construct a concrete perfect set of operators. In
Section~\ref{sparseMC}, we carry out such a construction for a set of
Markov equivalence\vadjust{\goodbreak} classes with sparsity constraints and provide
algorithms to obtain a reversible Markov chain.
We now show that a reversible Markov chain can be used to compute
interesting properties of a completed PDAG set $\cal S$.

\subsection{Estimating the properties of $\cal S$ by a perfect Markov chain}
\label{estimates}

For any ${\cal C}\in{\cal S}$,
let $f(\cal C)$ be a real function describing any property of interest
of~$\cal C$, and the random variable $u$ be uniformly distributed on
${\cal S}$. In order to understand the property of interest, we compute
the distribution of $f(u)$.

Let's consider one example in the literature. The proportion of Markov
equivalence classes of size one (equivalently, completed PDAGs that are
directed) in ${\cal S}_p$ is studied in the literature \cite
{gillispie2006formulas,gillispie2002size,pena2007approximate}. For this
purpose, we can define $f(u)$ as the size of Markov equivalence classes
represented by $u$ and obtain the proportion by computing the
probability of {$\{f(u)=1\}$.

Let $A$ be any subset of $\mathbb R$, the probability of $\{f(u)\in A\}
$ is
%
\begin{equation}
\label{peg} {\mathbb P} \bigl( f(u)\in A \bigr)= \frac{|\{{\cal
C}\dvtx f({\cal C})\in
A, {\cal C} \in{\cal S}\}|}{|{\cal S}|}=
\frac{\sum_{{\cal C}\in{\cal
S}}I_{\{f({\cal C})\in A\}}}{|{\cal S}|},
\end{equation}
where $ |{\cal S} |$ is the number of elements in the set
${\cal S}$ and $I$ is an indicator function.

Let $\{{e}_t\}_{t=1,\ldots,N}$ be a realization of Markov chain $\{e_t\}
$ on ${\cal S}$ based on a perfect operator set $\cal O$ according to
Definition~\ref{markovchain} and $M_t=M({\cal O}_{{e}_t})$. Let $\pi
({e}_t)$ be the stationary probability of Markov chain $\{e_t\}$. From
Proposition~\ref{stationaryd}, we have $\pi({e}_t)\propto M_t$ for
$t=1,\ldots,N$. We can use $\{{e}_t,M_t\}_{t=1,\ldots,N}$ to estimate
the probability of $\{ f(u)\in A \}$ by
%
\begin{equation}
\label{esteg} \hat{\mathbb P}_N \bigl(f(u)\in A \bigr)=
\frac{\sum_{t=1}^N I_{\{
f({e}_t)\in A\}}{M_t^{-1}}}{\sum_{t=1}^N {M_t^{-1}}}.
\end{equation}

From the ergodic theory of Markov chains (see Theorem 1.10.2 in \cite
{norris1997markov}), we can get Proposition~\ref{estimator} directly.

\begin{proposition}\label{estimator}
Let $\cal S$ be a given set of completed PDAGs, and assume the set of
operators ${\cal O}$ on $\cal S$ is perfect. The Markov chain $\{e_t\}
_{t=1,\ldots,N}$ is obtained according to Definition~\ref{markovchain}.
Then the estimator $\hat{\mathbb P}_N (\{f(u)\in A\} )$ in
equation \eqref{esteg} converges to
${\mathbb P} (\{f(u)\in A\} )$ in equation \eqref{peg} with
probability one, that is,
%
\begin{equation}
\label{estegth} \mathbb P \bigl( \hat{\mathbb P}_N \bigl(f(u)\in A
\bigr)\to{\mathbb P} \bigl( f(u)\in A \bigr) \mbox{ as } N\to\infty\bigr)=1.
\end{equation}
\end{proposition}

Proposition~\ref{estimator} shows that the estimator defined in
equation \eqref{esteg} is a consistent estimator of ${\mathbb P} (
f(u)\in A )$. We can study any given subset of Markov equivalence
classes via equation \eqref{esteg} if we can obtain $\{{e}_t\}
_{t=1,\ldots,N}$ and $\{M_t\}_{t=1,\ldots,N}$.
We now turn to construct a concrete perfect set of operators for a set
of completed PDAGs with sparsity constraints and then introduce
algorithms to run a reversible Markov chain.

\section{A Reversible Markov chain on completed PDAGs with sparsity constraints}
\label{sparseMC}

We define a set of Markov equivalence classes ${\cal S}_p^n$ with $p$
vertices and at most $n$ edges as follows:
%
\begin{equation}\label{edgec}
{\cal S}_p^n = \{{\cal C}\dvtx  {\cal C} \mbox{ is a
completed PDAG with }p \mbox{ vertices and } n_{\scriptscriptstyle{\cal
C} }\leq n \},
\end{equation}
where $n_{\scriptscriptstyle{\cal C}}$ is the number of edges in $\cal
C$. Recall that ${\cal S}_p$ denotes the set of all completed PDAGs
with $p$ vertices. Clearly, ${\cal S}_p^n={\cal S}_p$ when $n\ge p(p-1)/2$.

We now construct a perfect set of operators on ${\cal S}_p^n$. Notice
that our constructions can be extended to adapt to some other sets of
completed PDAGs, say, a set of completed PDAGS with a given maximum degree.
In Section~\ref{srandomwalk}, we construct the perfect set of
operators for any completed PDAG in ${\cal S}_p^n$.
In Section~\ref{algorithms}, we propose algorithms and their
accelerated version for efficiently obtaining a Markov chain based on
the perfect set of operators.

\subsection{Construction of a perfect set of operators on ${\cal S}_p^n$}\label{srandomwalk}

In order to construct a perfect set of operators, we need to define the
set of operators on each completed PDAG in ${\cal S}_p^n$. Let $\cal C$
be a completed PDAG in ${\cal S}_p^n$. We consider six types of
operators on $\cal C$ that were introduced in Section~\ref{subs22}:
InsertU, DeleteU, InsertD, DeleteD, MakeV and RemoveV.
The operators on $\cal C$ with the same type but different modified
edges constitute a set of operators. We introduce six sets of operators
on $\cal C$ denoted by
$\mathit{InsertU}_{\cal C}$, $\mathit{DeleteU}_{\cal C}$, $\mathit{InsertD}_{\cal C}$,
$\mathit{DeleteD}_{\cal C}$, $\mathit{MakeV}_{\cal C}$ and $\mathit{RemoveV}_{\cal C}$ in
Definition~\ref{constrainedset}. In addition to the conditions that
guarantee validity, for each type of operators, we also introduce other
constraints to make sure that all operators are reversible.

First we explain some notation used in Definition \ref
{constrainedset}. Let $x$ and $y$ be any two distinct vertices in $\cal
C$. The neighbor set of $x$ denoted by $N_{x}$ consists of every vertex $y$
with $x - y$ in $\cal C$. The common neighbor set of $x$ and $y$ is
defined as $N_{xy}=N_{x}\cap N_{y}$. $x $ is a \textit{parent} of $y$
and $y$ is a \textit{child} of $x$ if $x \rightarrow y$ occurs in $\cal
C$. A vertex $u$ is a common child of $x$ and $y$ if $u$ is a child of
both $x$ and $y$. $\Pi_{x }$ represents the set of all parents of $x$.

\begin{definition}[(Six sets of operators on $\cal C$)]
\label{constrainedset}
Let $\cal C$ be a completed PDAG in ${\cal S}_p^n$
and $n_{\cal C}$ be the number of edges in $\cal C$. We introduce six
sets of operators on $\cal C$: $\mathit{InsertU}_{\cal C}$ $\mathit{DeleteU}_{\cal C}$,
$\mathit{InsertD}_{\cal C}$, $\mathit{DeleteD}_{\cal C}$, $\mathit{MakeV}_{\cal C}$ and
RemoveV$_{\cal C}$ as follows.

\begin{longlist}[(d)]
\item[(a)] \label{test} For any two vertices $x, y$ that are not
adjacent in $\cal C$, the operator ``InsertU $x-y$'' on $\cal C$ is in
$\mathit{InsertU}_{\cal C}$ if and only if \textbf{$(\mbox{\textit{iu}}_1)$} $n_{\cal
C} < n$; \textbf{$(\textit{iu}_2)$} ``InsertU $x-y$'' is valid; \textbf
{$(\mbox{\textit{iu}}_3)$} for any $u$ that is a common child of $x,y$ in
${\cal C} $, both $x\to u$ and $y\to u$ occur in the resulting
completed PDAG of ``InsertU $x-y$.''
\item[(b)] For any undirected edge $x-y$ in $\cal C$, the operator
``DeleteU $x-y$'' on $\cal C$ is in $\mathit{DeleteU}_{\cal C}$ if and only if
\textbf{$(\mbox{\textit{du}}_1)$} ``DeleteU $x-y$'' is valid.
\item[(c)] For any two vertices $x, y$ that are not adjacent in $\cal
C$, the operator ``InsertD $x\to y$'' on $\cal C$ is in $\mathit{InsertD}_{\cal
C}$ if and only if\vadjust{\goodbreak} \textbf{$(\mbox{\textit{{id}}}_1)$} $n_{\cal C} < n$;
\textbf{$(\mbox{\textit{id}}_2 )$} ``InsertD $x\to y$'' is valid;
\textbf{$(\mbox{\textit{id}}_3)$} for any $u$ that is a common child of
$x,y$ in ${\cal C}$, $y\to u$ occurs in the resulting completed PDAG of
``InsertD $x\to y$.''
\item[(d)] For any directed edge $x\to y$ in $\cal C$, operator
``DeleteD $x\to y$'' on $\cal C$ is in $\mathit{DeleteD}_{\cal C}$ if and only if
\textbf{$(\mbox{\textit{dd}}_1)$} ``DeleteD $x\to y$'' is valid; \textbf{$(
\mbox{\textit{dd}}_2)$} for any $v$ that is a parent of $y$ but not a parent
of $x$, directed edge $v\to y$ in ${\cal C} $ occurs in the resulting
completed PDAG of ``DeleteD $x\to y$.''
\item[(e)] For any subgraph $x-z-y$ in $\cal C$, the operator ``MakeV
$x\to z \leftarrow y$'' on $\cal C$ is in $\mathit{MakeV}_{\cal C}$ if and only
if \textbf{$(\mbox{\textit{mv}}_1)$} ``MakeV $x\to z \leftarrow y$'' is valid.
\item[(f)] For any $v$-structure $x\to z\leftarrow y$ of $\cal C$, the
operator ``RemoveV $x\to z \leftarrow y$'' on $\cal C$ is in
RemoveV$_{\cal C}$ if and only if $(\rv_1)$ $\Pi
_{x}=\Pi_{y}$; $(\rv_2)$ $\Pi_{x}\cup N_{xy}=\Pi
_{z}\setminus\{x,y\}$; $(\rv_3)$ every undirected
path between $x$ and $y$ contains a vertex in $N_{xy}$.
\end{longlist}
\end{definition}

Munteanu and Bendou~\cite{munteanu2001eq} discuss the constraints for
the first five types of operators such that each one can transform one
completed PDAG to another. Chickering~\cite{chickering2002learning}
introduces the necessary and sufficient conditions such that these five
types of operators are valid. We list the conditions introduced by
Chickering~\cite{chickering2002learning}
in Lemma~\ref{chickervalid}, Appendix~\ref{appA.1}, and employ them to guarantee
that the conditions {\textbf{iu}}$_2$, {\textbf{du}}$_1$, {\textbf
{id}}$_2$, {\textbf{dd}}$_1$ and {\textbf{mv}}$_1$ in Definition \ref
{constrainedset} hold.

The set of operators on $\cal C$ denoted by ${\cal O}_{\cal C}$ is
defined as follows:
%
\begin{eqnarray}
\label{constraintedC} {\cal O}_{\cal C} &=&\mathit{InsertU}_{\cal C} \cup
\mathit{DeleteU}_{\cal C}\cup \mathit{InsertD}_{\cal C}\nonumber\\[-8pt]\\[-8pt]
&&{}\cup \mathit{DeleteD}_{\cal C} \cup
\mathit{MakeV}_{\cal C} \cup \mathit{RemoveV}_{\cal C}.\nonumber
\end{eqnarray}

Taking the union over all completed PDAGs in ${\cal S}_p^n$, we define
the set of operators on ${\cal S}_p^n$ as
%
\begin{equation}
\label{perfectset} {\cal O} =\bigcup_{{\cal C} \in{\cal S}_p^n} {\cal
O}_{\cal C},
\end{equation}
where ${\cal O}_{\cal C}$ is the set of operators in equation \eqref
{constraintedC}.
In the main result of this paper, we show that ${\cal O}$ in equation
\eqref{perfectset} is a perfect set of operators on ${\cal S}_p^n$.

\begin{theorem}[(A perfect set of operators on ${\cal S}_p^n$)]
\label{perfect}
${\cal O}$ defined in equation \eqref{perfectset} is a perfect set of
operators on ${\cal S}_p^n$.
\end{theorem}

Here we notice that \textbf{iu}$_3$, \textbf{id}$_3$ and \textbf
{dd}$_2$ are key conditions in Definition~\ref{constrainedset} to
guarantee that $\cal O$ is reversible.
Without these three conditions, there are operators that are not
reversible; see Example 3, Section 2.1 in the Supplementary Material
\cite{hesupp}.
We provide a proof of Theorem~\ref{perfect} in Appendix~\ref{appA.2}.

The preceding section showed how to construct a perfect set of operators.
A~toy example is provided as Example 4 in Section 2.1 of the
Supplementary Material~\cite{hesupp}. Based on the perfect set of
operators we can obtain a finite irreducible reversible discrete-time chain.
In the next subsection, we provide detailed algorithms for obtaining a
Markov chain on ${\cal S}_p^n$ and their accelerated version.

\subsection{Algorithms}
\label{algorithms}

In this subsection, we provide the algorithms in detail to generate a
Markov chain on ${\cal S}_p^n $, defined in Definition \ref
{markovchain} based on the perfect set of operators defined in
\eqref{perfectset}. A sketch of Algorithm~\ref{mc} is shown below; some steps
of this algorithm are further explained in the subsequent algorithms.


\begin{algorithm}[t]
{\fontsize{9.5pt}{13pt}\selectfont{
\caption{Road map to construct a Markov chain on ${\cal S}_p^n$}
\label{mc}
\KwIn{\\$p$, the number of vertices;
$n$, the maximum number of edges;
$N$, the length of Markov chain.}
\KwOut{\\
$\{e_t,M_t\}_{t=1,\ldots,N}$, where $\{e_t\}$ is Markov chain
and $M_t$ is the number of operators in ${\cal O}_{e_t} $.}

Initialize $e_0$ as any completed PDAG in ${\cal S}_p^n$

\For{$t \leftarrow0$ \KwTo$N$}
{
\begin{description}
\item[Step A] Construct the set of operators ${\cal O}_{e_t}$ in
equation \eqref{constraintedC} via Algorithm~\ref{msetalg};
\item[Step B] Let $M_t$ be the number of operators in ${\cal O}_{e_t} $;
\item[Step C] Randomly choose an operator $o$ uniformly from ${\cal
O}_{e_t} $;
\item[Step D] Apply operator $o$ to $e_t$. Set $e_{t+1}$ as the
resulting completed PDAG of $o$.
\end{description}
}
\Return{$\{e_t,M_t\}_{t=1,\ldots,N}.$}}}
\end{algorithm}

Step A of Algorithm~\ref{mc} constructs the sets of operators on
completed PDAGs in the chain $\{e_t\}$. It is the most difficult step
and dominates the time complexity of Algorithm~\ref{mc}. Step B and
Step C can be implemented easily after ${\cal O}_{e_t}$ is obtained.
Step D can be implemented via Chickering's method \cite
{chickering2002learning} that was mentioned in Section~\ref{subs22}.
We will show that the time complexity of obtaining a Markov chain on
${\cal S}_p^n$ with length $N$ ($\{e_t\}_{t=1,\ldots,N}$) is
approximate $O(Np^3)$ if $n$ is the same order of $p$. For large $p$,
we also provide an accelerated version that, in some cases, can run
hundreds of times faster.

The rest of this section is arranged as follows. In Section
\ref{algorithmsa}, we first introduce the algorithms to implement Step A.
In Section~\ref{acceleration} we discuss the time complexity of our
algorithm, and provide an acceleration method to speed up Algorithm~\ref{mc}.

\subsubsection{\texorpdfstring{Implementation of Step A in Algorithm \protect\ref{mc}}
{Implementation of Step A in Algorithm 1}}\label{algorithmsa}

A detailed implementation of Step A (to construct ${\cal O}_{e_t}$)
is described in Algorithm~\ref{msetalg}. To construct ${\cal O}_{e_t}$
in Algorithm~\ref{msetalg}, we go through all possible operators on
$e_t$ and
choose those satisfying the corresponding conditions in Definition \ref
{constrainedset}.

The conditions in Algorithm~\ref{msetalg} include: \textbf{iu}$_1$,
\textbf{iu}$_2$, \textbf{iu}$_3$, \textbf{du}$_{1}$, \textbf{id}$_1$,
\textbf{id}$_{2}$, \textbf{id}$_3$, \textbf{dd}$_{1}$, \textbf{dd}$_2$,
\textbf{rm}$_1$, \textbf{rv}$_1$, \textbf{rv}$_2$ and \textbf{mv}$_{1}$.
For each possible operator, we check the corresponding conditions
shown in Algorithm~\ref{msetalg} one-by-one until one of them fails.
Below, we introduce how to check these conditions.

\renewcommand{\thealgocf}{1.1}
\begin{algorithm}[t]
{\fontsize{10.4pt}{13pt}\selectfont{
\caption{Construct ${\cal O}_{e_t}$ for a completed PDAG $e_t$.}
\label{msetalg}
\KwIn{A completed PDAG ${e_t}$ with $p$ vertices.
}
\KwOut{Operator set ${\cal O}_{e_t}$.\\
{{\fontsize{9.4pt}{13pt}\selectfont{\texttt{// All sets of possible modified edges of $e_t$ used below,\\
\hspace*{15pt}for example, Undirected-edges$_{e_t}$, are generated according\\
\hspace*{15pt}to Definition
\ref{constrainedset}.}}}}
}
Set ${\cal O}_{e_t}$ as empty set\\
\For{each undirected edge $x-y$ in Undirected-edges$_{e_t}$}
{consider operator DeleteU $x -x $, add it to ${\cal O}_{e_t}$ if \textbf
{du}$_{1}$ holds,}
\For{each directed edge $x\to y$ in Directed{-}edges$_{e_t}$}
{consider DeleteD $x \to x $, add it to ${\cal O}_{e_t}$ if both \textbf
{dd}$_{1}$ and \textbf{dd}$_2$ hold;}
\For{ each $v$-structure $x\to z\leftarrow y$ in V{-}structures$_{e_t}$ }
{consider RemoveV $x_k\to x_i\leftarrow x_l$, add it to ${\cal
O}_{e_t}$ if \textbf{rv}$_1$, \textbf{rv}$_2$ and \textbf{rv}$_3$ hold,
}
\For{each undirected $v$-structure $x-z-y$ in
Undirected{-}$v${-}structures$_{e_t}$ }
{consider MakeV $x_k\to x_i\leftarrow x_l$, add it to ${\cal O}_{e_t}$
if \textbf{mv}$_{1}$ holds,
}
\If{$n_{e_t}<n$ $($i.e., \textbf{iu}$_1$ or \textbf{id}$_1$ holds$)$}
{
\For{ each pair $(x,y)$ in Pairs{-}nonadj$_{e_t}$}
{consider InsertU $x-y$, add it to ${\cal O}_{e_t}$ if \textbf
{iu}$_1$, \textbf{iu}$_{2}$, and \textbf{iu}$_3$ hold; \\
consider InsertD $x\to y$, add it to ${\cal O}_{e_t}$, if \textbf
{id}$_1$, \textbf{id}$_{2}$ and \textbf{id}$_3$ hold; \\
consider InsertD $x\leftarrow y$, add it to ${\cal O}_{e_t}$ if \textbf
{id}$_1$, \textbf{id}$_{2}$ and \textbf{id}$_3$ hold.
}
}
\Return{${\cal O}_{e_t}$}}}
\end{algorithm}

The conditions \textbf{iu}$_3$, \textbf{id}$_3$ and \textbf{dd}$_2$ in
Algorithm~\ref{msetalg} depend on both $e_t$ and the resulting
completed PDAGs of the operators. Intuitively, checking \textbf
{iu}$_3$, \textbf{id}$_3$ or \textbf{dd}$_2$ requires that we obtain
the corresponding resulting completed PDAGs. We know that the time
complexity of getting a resulting completed PDAG of $e_t$ is
$O(pn_{e_t})$~\cite{dor1992simple,chickering2002learning}, where
$n_{e_t}$ is the number of edges in $e_t$. To avoid generating
resulting completed PDAG, in the Supplementary Material~\cite{hesupp},
we provide three algorithms to check \textbf{iu}$_3$, \textbf{id}$_3$
and \textbf{dd}$_2$ only based on $e_t$ and in an efficient manner.

The other conditions can be tested via classical graph algorithms.
These tests include:
(1) whether two vertex sets are equal or not, (2) whether a subgraph
is a clique or not and (3) whether all partially directed paths or all
undirected paths between two vertices contain at least one vertex in a
given set. Checking the first two types of conditions is trivial and
very efficient because the sets involved are small for most completed
PDAGs in ${\cal S}_p^n$ when $n$ is of the same order of $p$. To check
the conditions with the third type, we just need to check whether there
is a partially directed path or undirected path between two given
vertices not through any vertices in the given set. We check this using
a depth-first search from the source vertex. When looking for an
undirected path, we can search within the corresponding chain component
that includes both the source and the destination vertices.

\subsubsection{\texorpdfstring{Time complexity of Algorithm \protect\ref{mc} and an accelerated version}
{Time complexity of Algorithm 1 and an accelerated version}}
\label{acceleration}

We now discuss the time complexity of Algorithm~\ref{mc}. For $e_t \in
{\cal S}_p^n$, let $p$ and $n_t$ be the number of vertices and edges in
$e_t$, respectively, $k_t$ be the number of $v$-structures in~$e_t$, and
$k'_t$ be the number of undirected $v$-structures (subgraphs $x-y-z$ with
$x$ and $z$ nonadjacent) in $e_t$. To construct ${\cal O}_{e_t}$, in
Step A of Algorithm~\ref{mc} (equivalently, Algorithm~\ref{msetalg}),
all possible operators we need to go through: $n_t$ deleting operators
(DeleteU and DeleteD), $3(p(p-1)/2-n_t)$ inserting operators (InsertU
and InsertD) when the number of edges in $e_t$ is less than $n$, $k_t$
RemoveV operators and $k'_t$ MakeV operators. There are at most
$Q_t=1.5p(p-1)-2n_t+k_t+k'_t$ possible operators for $e_t$. Among all
conditions in Algorithm~\ref{msetalg}, the most time-consuming one,
which takes time $O(p+n_t)$~\cite{chickering2002learning}, is to look
for a path via the depth-first search for an operator with type of InsertD.
We have that the time complexity of constructing ${\cal O}_{e_t}$ in
Algorithm~\ref{msetalg} is $O(Q_t(p+n_t))$ in the worst case and the
time complexity of Algorithm~\ref{mc} is $O({\sum_{t=1}^N}Q_t(p+n_t))$
in the worst case, where $N$ is the length of Markov chain in Algorithm
\ref{mc}. We know that $k_t $ and $k'_t$ reach the maxima
$(p-2)/2*\mbox{floor}(p/2)*\mbox{ceil}(p/2)$ when $e_t$ is a evenly divided complete
bipartite graphs~\cite{gillispie2002size}. Consequently, the time
complexity of Algorithm~\ref{mc} are $O(Np^4)$ in the worst case.
Fortunately, when $n$ is a few times of $p$, say $n=2p$,
all completed PDAGs in ${\cal S}_p^n$ are sparse and our experiments
show $k_t$ and $k'_t$ are much less than $O(p^2)$ for most completed
PDAGs in Markov chain $\{e_t\}_{t=1,\ldots,N}$. Hence the time
complexity of Algorithm~\ref{mc} is approximate $O(Np^3)$ on average
when $n$ is a few times of $p$.

We can implement Algorithm~\ref{mc} efficiently when $p$ is not large
(less or around 100 in our experiments). However, when $p$ is larger,
we need large $N$ to guarantee the estimates reach convergence.
Experiments in Section~\ref{experiments} show $N=10^6$ is suitable. In
this case, cubic complexity ($O(Np^3)$) of Algorithm~\ref{mc} is
unacceptable. We need to speed up the algorithms for a very large $p$.

Notice that in Algorithm~\ref{mc}, we obtain an irreducible and
reversible Markov chain $\{e_t\}$ and a sequence of numbers $\{M_t\}$
by checking all possible operators on each $e_t$. The sequence $\{M_t\}
$ are used to compute the stationary probabilities of $\{e_t\}$
according to Proposition~\ref{stationaryd}. We now introduce an
accelerated version of Algorithm~\ref{mc} to generate irreducible and
reversible Markov chains on ${\cal S}_p^n$. The basic idea is that we
do not check all possible operators but check some random samples.
These random samples are then used to estimate $\{M_t\}$.

We first explain some notation used in the accelerated version. For
each completed PDAG $e_t$, if $n_{e_t}<n$, ${\cal O}_{e_t}^{(\mathrm{all})}$ is
the set of all possible operators on $e_t$ with types of InsertU,
DeleteU, InsertD, DeleteD,\vadjust{\goodbreak} MakeV and RemoveV. If $n_{e_t}=n$, the
number of edges in $e_t$ reaches the upper bound $n$, no more edges can
be inserted into $e_t$. Let ${\cal O}_{e_t}^{(-\mathrm{insert})}$ be the set of
operators obtained by removing operators with types of InsertU and
InsertD from ${\cal O}_{e_t}^{(\mathrm{all})}$. ${\cal O}_{e_t}^{(-\mathrm{insert})}$ is
the set of all possible operators on $e_t$ when $n_{e_t}=n$. We can
obtain ${\cal O}_{e_t}^{(\mathrm{all})}$ and ${\cal O}_{e_t}^{(-\mathrm{insert})}$ easily
via all possible modified edges introduced in Algorithm~\ref{msetalg}.
The accelerated version of Algorithm~\ref{mc} is shown in Algorithm~\ref{acc}.

%

\renewcommand{\thealgocf}{2}
\begin{algorithm}[t]
{\fontsize{10.2pt}{13pt}\selectfont{
\caption{An accelerated version of Algorithm \protect\ref{mc}.}
\label{acc}
\SetKwBlock{Block}{Step}{end}
\KwIn{\\$\alpha\in(0,1]$: an acceleration parameter; $p$, $n$ and $N$,
the same as input in Algorithm~\ref{mc}}
\KwOut{\\
$\{e_t,\hat{M}_t\}_{t=1,\ldots,N}$, where
$\hat{M}_t$ is an estimation of $M_t=|{\cal O}_{e_t}|$}

Initialize $e_0$ as any completed PDAG in ${\cal S}_p^n$

\For{$t \leftarrow0$ \KwTo$N$}
{
\Block(\textbf{A$'$:}){
\If{$n_{e_t}<n$ }{Set ${\cal O}'_{e_t}={\cal O}_{e_t}^{(\mathrm{all})}$ }
\Else{Set ${\cal O}'_{e_t}={\cal O}_{e_t}^{(-\mathrm{insert})}$}
Set $m_t=|{\cal O}'_{e_t}|$\\
Randomly sample $[\alpha m_t]$ operators without replacement from
${\cal O}'_{e_t}$ to generate a set ${\cal O}_{e_t}^{(\mathrm{check})}$, where
$[\alpha m_t]$ is the integer closest to $\alpha m_t$. \\\label{line7}
Check all operators in ${\cal O}_{e_t}^{(\mathrm{check})}$, and choose perfect
operators from it to construct a set of operators $\tilde{\cal
O}_{e_t}$. \\
Set $m_{_t}^{_{_{(\tilde{\cal O})}}}=|\tilde{\cal O}_{e_t}|$. If
$m_{_t}^{_{_{(\tilde{\cal O})}}}=0$, go to line 9.}
\Block(\textbf{ B$'$:}){ Let
$\hat{M}_t=m_{t}\frac{m_{_t}^{_{_{(\tilde{\cal O})}}}}{[\alpha m_t]},$}
\Block(\textbf{ C$'$:}){ Randomly choose an operator $o$ uniformly
from $\tilde{\cal O}_{e_t}$.
}
\Block(\textbf{ D:}){ Apply operator $o$ to $e_t$. Set $e_{t+1}$ as
the resulting completed PDAG of $o$.
}
}
\Return{$\{e_t,\hat{M}_t\}_{t=1,\ldots,N}.$}}}
\end{algorithm}

In Algorithm~\ref{acc}, ${\cal O}'_{e_t}$ (either ${\cal
O}_{e_t}^{(\mathrm{all})}$ or ${\cal O}_{e_t}^{(-\mathrm{insert})}$) is the set of all
possible operators on $e_t$, $\alpha\in(0,1]$ is an acceleration
parameter that determines how many operators in ${\cal O}'_{e_t}$ are
checked, ${\cal O}_{e_t}^{(\mathrm{check})}$ is a set of checked operators that
are randomly sampled without replacement from ${\cal O}'_{e_t}$ and
$\tilde{\cal O}_{e_t}$ is the set of all perfect operators in ${\cal
O}_{e_t}^{(\mathrm{check})}$. When $\alpha=1$, $\tilde{\cal O}_{e_t}={\cal
O}_{e_t}$ and Algorithm~\ref{acc} becomes back to Algorithm~\ref{mc}.

In Algorithm~\ref{acc}, because the operators in $\tilde{\cal
O}_{e_t}$ are i.i.d. sampled from ${\cal O}_{e_t}$ in Step~A$'$ and
operator $o$ is chosen uniformly from $\tilde{\cal O}_{e_t}$ in Step
C$'$, clearly, $o$ is also chosen uniformly from ${\cal O}_{e_t}$. We
have that the following Corollary~\ref{accstationaryd} holds according
to Proposition~\ref{stationaryd}.

\begin{corollary}[(Stationary distribution of $\{e_t\}$ on ${\cal
S}_p^n$)]\label{accstationaryd}
Let ${\cal S}_p^n$, defined in equation \eqref{edgec}, be the set of
completed PDAGs with $p$ vertices and maximum $n$ of edges, ${\cal
O}_{e_t}$, defined in equation \eqref{constraintedC}, be the set of
operators on $e_t$, and $M_t$ be the number of operators in ${\cal O}_{e_t}$.
For the Markov chain $\{e_t\}$ on ${\cal S}_p^n$ obtained via
Algorithms~\ref{mc} or~\ref{acc}, then:
\begin{longlist}[(2)]
\item[(1)] the Markov chain $\{e_t\}$ is irreducible and reversible;
\item[(2)] the Markov chain $\{e_t\}$ has a unique stationary
distribution $\pi$ and $\pi(e_t)\propto M_t$.
\end{longlist}
\end{corollary}

In Algorithm~\ref{acc}, we provide an estimate of $M_t$ instead of
calculating it exactly in Algorithm~\ref{mc}. Let $|{\cal
O}'_{e_t}|=m_t$, $|{\cal O}_{e_t}^{(\mathrm{check})}|=[\alpha m_t]$ and $|\tilde
{\cal O}_{e_t}|=m_{_t}^{_{_{(\tilde{\cal O})}}}$. Clearly, the ratio
${m_{_t}^{_{_{(\tilde{\cal O})}}} / [\alpha m_t]}$ is an unbiased
estimator of the population proportion $M_t/m_t$ via sampling without
replacement.
We can estimate ${M}_t=|{\cal O}_{e_t}|$ in Step B$'$ as
%
\begin{equation}
\label{speedingformula}
\hat{M}_t=m_{t}\frac{m_{_t}^{_{_{(\tilde{\cal O})}}}}{[\alpha m_t]}.
\end{equation}

We have that when $ [\alpha m_t]$ is large, the estimator $\hat{M}_t$
has an approximate normal distribution with mean equal to $M_t=|{\cal
O}_{e_t}|$.

Let the random variable $u$ be uniformly distributed on ${\cal
S}_p^n$, $f(u)$ be a real function describing a property of interest of
$u$ and $A$ be a subset of $\mathbb{R}$. By replacing $M_t$ with $\hat
{M}_t$ in equation \eqref{esteg}, we estimate ${\mathbb P}_N (\{
f(u)\in A\} )$ via $\{e_t,\hat{M}_t\}_{t=1,\ldots,N}$ as follows:
%
\begin{equation}
\label{estegspeed} \hat{\mathbb P}_N' \bigl(f(u)\in A
\bigr)= \frac{\sum_{t=1}^N I_{\{
f(e_t)\in A\}}\hat{M}_t^{-1}}{\sum_{t=1}^N\hat{M}_t^{-1}},
\end{equation}
where ${\mathbb P}_N (f(u)\in A )$ is defined in equation
\eqref{peg}.

In the accelerated version, only $100\alpha\%$ of all possible operators
on $e_t$ are checked. In Section~\ref{experiments}, our experiments on
${\cal S}_{100}^{150} $ show that
the accelerated version can speed up the approach nearly $\frac
{1}{\alpha}$ times, and that
equation \eqref{estegspeed} provides almost the same results as
equation \eqref{esteg}
in which $\{e_t,{M}_t\}_{t=1,\ldots,N}$ from Algorithm~\ref{mc} are
used. Roughly speaking, if we set $\alpha=1/p$, the time complexity of
our accelerated version can
reduce to $O(Np^2)$.

\section{Experiments}
\label{experiments}

In this section, we conduct experiments to illustrate
the reversible Markov chains proposed in this paper and their
applications for studying Markov equivalence classes. The main points
obtained from these experiments are as follows:
\begin{longlist}[(2)]
\item[(1)] For ${\cal S}_p$ with small $p$, the estimations of our
proposed are very close to true values. For ${\cal S}_p^n$ with large
$p$ (up to 1000), the accelerated version of our proposed approach is
also very efficient, and the estimations in equations \eqref{esteg} and
\eqref{estegspeed} converge quickly as the length of Markov chain increases.


\item[(2)] For
completed PDAGs in ${\cal S}_p^n$ with sparsity constraints ($n$ is a
small multiple of $p$), we see that (i) most edges are directed, (ii)
the sizes of maximum chain components
(measured by the number of vertices) are very small (around ten) even
for large $p$ (around 1000) and (iii) the number
of chain components
grows approximately linearly with $p$.
\end{longlist}

As we know, under the assumption that there are no latent or selection
variables present, causal inference based on observational data will
give a completed PDAG. Interventions are needed to infer the directions
of the undirected edges in the completed PDAG. Our results show that if
the underlying completed PDAG is sparse, in the model space of Markov
equivalence classes, most graphs have few undirected edges and small
chain components. They give hope for learning causal relationships via
observational data and for inferring the directions of the undirected
edges via interventions.

In Section~\ref{exsize}, we evaluate our methods by comparing the size
distributions of Markov equivalence classes in ${\cal S}_p$ with small
$p$ to true distributions ($p=3,4$) or Gillispie's results ($p=6$) \cite
{gillispie2002size}. In Section~\ref{sparseconstraints}, we report the
proportion of directed edges and the properties of chain components of
Markov equivalence classes under sparsity constraints. In Section~\ref{accsimu},
we show experimentally that Algorithm~\ref{acc} is much faster than
Algorithm~\ref{mc}, and that the difference in the estimates obtained
is small.
Finally, we study the asymptotic properties of our proposed estimators
in Section~\ref{exmixing}.


\subsection{Size distributions of Markov equivalence classes in ${\cal S}_p$ for small $p$}
\label{exsize}

We consider size distributions of completed PDAGs in ${\cal S}_p$ for
$p=3,4$ and~$6$, respectively.
There are 11 Markov equivalence classes in ${\cal S}_3$, and 185
Markov equivalence classes in ${\cal S}_4$. Here we can get the true
size distributions for ${\cal S}_3$ and ${\cal S}_4$ by listing all the
Markov equivalence classes and calculating the size of each explicitly.
Gillespie and Perlman calculate the true size probabilities for ${\cal
S}_6$ by listing all classes; these are denoted as GP-values. We
estimate the size probabilities via equation\vadjust{\goodbreak} \eqref{esteg} with the
Markov chains from Algorithm~\ref{mc}. We ran ten independent Markov
chains using Algorithm~\ref{mc} to calculate the mean and standard
deviation of each estimate.
The results are shown in Table~\ref{p610}, where $N$ is the sample size
(length of Markov chain). We can see that the means are very close to
true values or GP-values, and the standard deviations are also very small.

\begin{table}
\tablewidth=315pt
\caption{Size distributions for ${\cal S}_p$ with $p=3,4$ and $6$,
respectively. $N$ is the sample size, $T$ is the time (seconds) used to
estimate the size distributions with a Markov chain, GP-values are
obtained by Gillispie and Perlman~\cite{gillispie2002size}}
\label{p610}
\begin{tabular}{@{}cc@{}}
\begin{tabular}{@{}ld{1.6}c@{}}
\hline
\multicolumn{3}{@{}l}{$\bolds{p=3}$\textbf{,} $\bolds{N=10^4}$\textbf{,} $\bolds{T=2}$ \textbf{sec}}\\
\hline
\textbf{Size}& \multicolumn{1}{c}{\textbf{True value}} & \textbf{Mean (Std)} \\
\hline
1& 0.36363^* & 0.36422 (0.00540) \\
2&0.27273 & 0.27160 (0.00412) \\
3&0.27273 & 0.27274 (0.00217) \\
6& 0.0909 & 0.09144 (0.00262) \\
\\\\\\\\
\end{tabular}\qquad
\begin{tabular}{@{}ld{1.6}c@{}}
\hline
\multicolumn{3}{@{}l}{$\bolds{p=4}$\textbf{,} $\bolds{N=10^4}$\textbf{,}
$\bolds{T=3}$ \textbf{sec}}\\
\hline
\textbf{Size}& \multicolumn{1}{c}{\textbf{True value}} & \textbf{Mean (Std)} \\
\hline
\hphantom{0}1&0.31892^*& 0.31859 (0.00946) \\
\hphantom{0}2& 0.25946& 0.25929 (0.00590) \\
\hphantom{0}3& 0.19460& 0.19572 (0.00635) \\
\hphantom{0}4& 0.10270 & 0.10229 (0.00395) \\
\hphantom{0}6& 0.02162 & 0.02162 (0.00145) \\
\hphantom{0}8& 0.06486 & 0.06464 (0.00291) \\
10&0.03243& 0.03249 (0.00202) \\
24&0.00540 & 0.00536 (0.00078) \\
\end{tabular}
\end{tabular}
\\
\hspace*{-7pt}\begin{tabular}{@{}ld{1.6}c@{\quad\qquad}ccc@{}}
\hline
\multicolumn{3}{@{}l}{$\bolds{p=6}$\textbf{,} $\bolds{N=10^5}$\textbf{,}
$\bolds{T=60}$ \textbf{sec}}\\
\hline
\textbf{Size}& \multicolumn{1}{c}{\textbf{GP-value}} & \textbf{Mean (Std)}
& \textbf{Size} & \textbf{GP-value} & \textbf{Mean (Std)} \\
\hline
\hphantom{0}1& 0.28667^* & 0.28588 (0.00393) & \hphantom{0}48&0.00013& 0.00013 (0.00004) \\
\hphantom{0}2&0.25858& 0.25897 (0.00299) & \hphantom{0}50&0.00034& 0.00034 (0.00007) \\
\hphantom{0}3&0.17064& 0.17078 (0.00248) & \hphantom{0}52&0.00017& 0.00018 (0.00003) \\

%
\multicolumn{3}{c}{$\vdots$} & \hphantom{0}54&0.00017& 0.00018 (0.00004) \\
28&0.00017& 0.00017 (0.00004) & \hphantom{0}60&0.00019& 0.00020 (0.00004) \\
30&0.00169& 0.00170 (0.00017) & \hphantom{0}72 &0.00006& 0.00006 (0.00002) \\
32&0.00236& 0.00238 (0.00017) & \hphantom{0}88 &0.00004& 0.00004 (0.00001) \\
36&0.00052& 0.00053 (0.00008) & 144&0.00009& 0.00009 (0.00003) \\
38&0.00034& 0.00035 (0.00004) & 156&0.00006& 0.00006 (0.00003) \\
40&0.00118& 0.00120 (0.00010) & 216 &0.00001& 0.00001 (0.00002) \\
42&0.00051& 0.00052 (0.00009) & \\
\hline
\end{tabular}
\end{table}

We implemented our proposed method (Algorithm~\ref{mc}, the version without
acceleration) in Python, and ran it on a computer with a 2.6 GHZ
processor. In Table~\ref{p610}, $T$ is the time used to estimate the
size distribution for ${\cal S}_3 $, ${\cal S}_4 $ or ${\cal S}_6 $.
These results were obtained within at most tens of seconds. In
comparison, a MCMC method in~\cite{pena2013approximate} took more than
one hour (in C$++$ on a 2.6 GHZ computer) in order to get similar
estimates of the proportions of Markov equivalence classes
of size one. It is worth noting that our estimates are based on a
single Markov chain, while the results in~\cite{pena2013approximate}
are based on $10^4$ independent Markov chains with $10^6$
steps.\vadjust{\goodbreak}

\subsection{Markov equivalence classes with sparsity constraints}\label
{sparseconstraints}

We now study the sets ${\cal S}_{p}^{n}$ of Markov equivalence classes
defined in equation \eqref{edgec}. The number of vertices $p$ is set to
$100,200,500$ or $1000$, and the maximum edge constraint $n$ is set to
$rp$ where $r$ is the ratio of $n$ to $p$. For each $p$, we consider
three ratios: 1.2, 1.5 and 3. The completed PDAGs in ${\cal
S}_{p}^{rp}$ are sparse since $r\leq3$. Define the size of a chain
component as the number of vertices it contains. In this section, we
report four distributions for completed PDAGs in ${\cal S}_{p}^{rp}$:
the distribution of proportions of directed edges, the distribution of
the numbers of chain components and the distribution of the maximum
size of chain components. The results about the distribution of the
numbers of $v$-structures are reported in the Supplementary Material~\cite
{hesupp}. In each simulation, given $p$ and $r$, a Markov chain with
length of $10^6$ on ${\cal S}_{p}^{rp}$ is generated via Algorithm \ref
{acc} to estimate the distributions via equation \eqref{estegspeed}.
The acceleration parameter $\alpha$ is set to $0.1,0.05,0.01$ and $
0.001$ for $p=100,200,500$ and $1000$, respectively.

In Figure~\ref{und}, twelve distributions of proportions of directed
edges are reported for ${\cal S}_{p}^{rp}$ with different $p$ and ratio
$r$. We mark the minimums, $5\%$ quartiles (solid circles below boxes),
1st quartiles, medians, 3rd quartiles and maximums of these distributions.
We can see that for a fixed $p$, the proportion of directed edges
increases with the number of edges in the completed PDAG. For example,
when the ratio $r=1.2$, the medians (red lines in boxes) of proportions
are near $92\%$; when the ratio $r=1.5$, the medians are near $95\%$;
when ratio $r=3$, the medians are near $98$\%.

\begin{figure}

\includegraphics{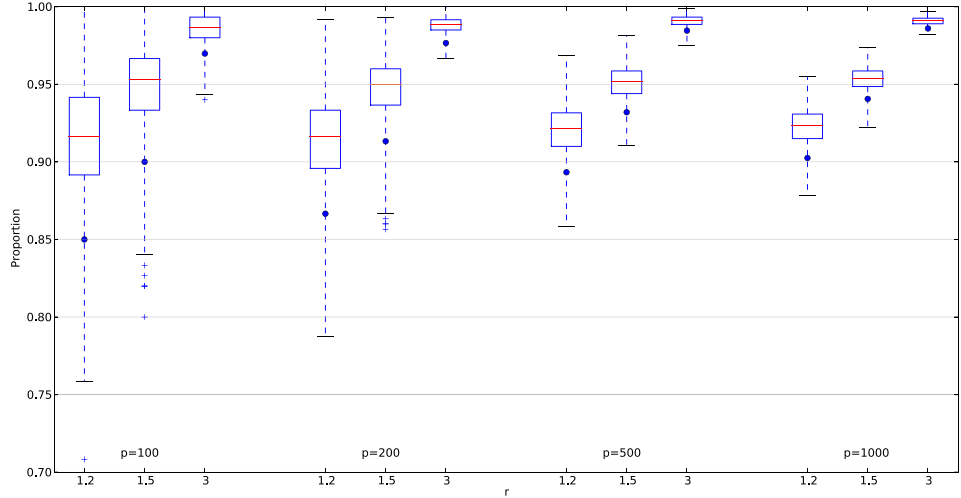}

\caption{Distribution of proportion of directed edges in completed
PDAGs in ${\cal S}_p^{rp}$. The lines in the boxes and the solid
circles under the boxes indicate the medians and the 5$\%$ quartiles,
respectively.} \label{und}
\end{figure}

%
\begin{figure}

\includegraphics{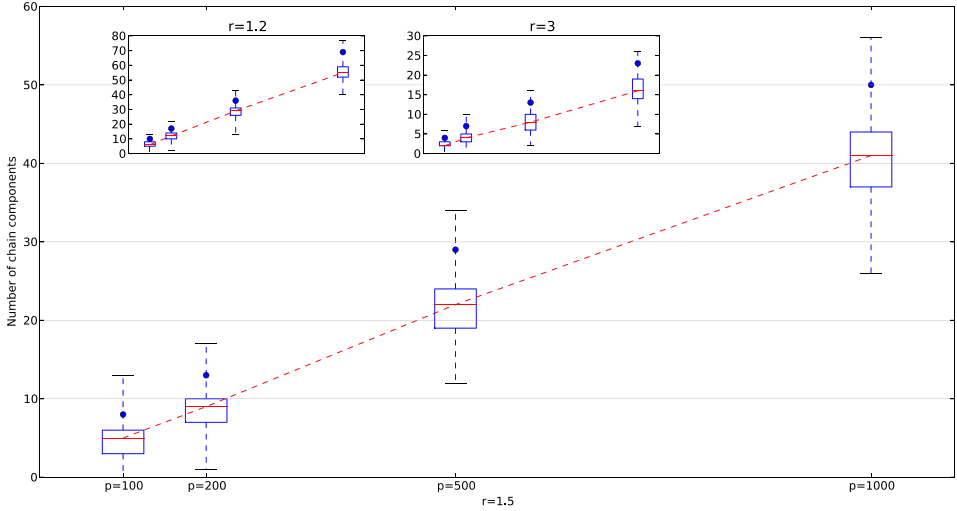}

\caption{Distributions of numbers of chain components of completed
PDAGs in $\mathcal{S}_p^{rp}$. The lines in the boxes and the solid
circles above the boxes indicate the medians and the 95$\%$ quartiles,
respectively.} \label{boxncc}
\end{figure}

%

The distributions of the numbers of chain components of completed PDAGs
in ${\cal S}_p^{rp}$ are shown in Figure~\ref{boxncc}. We plot the
distributions for ${\cal S}_p^{1.5p}$ in the main window and the
distributions for $r=1.2$ and $r=3$ in two sub-windows. We can see that
the medians
of the numbers of chain components are close to 5, 10, 20, and 40 for
completed PDAGs in
$\mathcal{S}_p^{1.5p}$ with $p=100,200,500$ and $1000$,
respectively. It seems that
there is a linear relationship between the number of chain components
and the number of
vertices $p$. In the insets, similar results are shown in the
distributions for $r=1.2$ and $r=3$.

The distributions of the maximum sizes of chain components of completed
PDAGs in $\mathcal{S}_p^{rp}$ are shown in Figure~\ref{Mnncc}. For
$\mathcal{S}_p^{1.5p}$ in the main window, the medians of the four
distributions are approximately 4, 5, 6 and 7 for $p=100, 200, 500$ and
1000, respectively. This shows that the maximum size of chain components
in a competed PDAG increases very slowly with $p$. In particular, from
the 95$\%$ quartiles (solid circles above boxes), we can see that the
maximum chain components of more than $95\%$ completed PDAGs in
$\mathcal{S}_p^{1.5p}$ have at most 8, 9, 10 and 13 vertices for
$p=100,200,500$ and $1000$, respectively. This result implies that sizes
of chain components in most sparse completed PDAGs are small.

\begin{figure}

\includegraphics{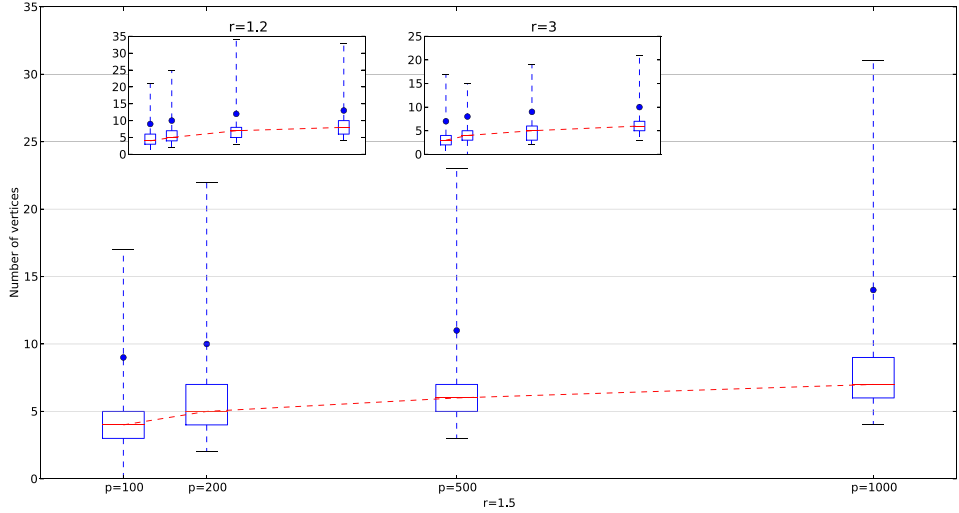}

\caption{The distributions of the maximum sizes of chain components of
completed PDAGs in $\mathcal{S}_p^{rp}$. The lines in the boxes and the
solid circles above the boxes indicate the medians and the 95$\%$
quartiles, respectively.} \label{Mnncc}
\end{figure}

%
%

\subsection{\texorpdfstring{Comparisons between Algorithm \protect\ref{mc} and its accelerated version}
{Comparisons between Algorithm 1 and its accelerated version}} \label{accsimu}

In this section, we show experimentally that the accelerated version
Algorithm~\ref{acc} is much faster than Algorithm~\ref{mc}, and the
difference of estimates based on two algorithms is small.
We have estimated four distributions on ${\cal S}_{100}^{150}$
in\vadjust{\goodbreak}
Section~\ref{sparseconstraints} via Algorithm~\ref{acc}. The four
distributions are
the distribution of proportions of directed edges, the distribution of
the numbers of chain components, the distribution of maximum size of
chain components and the distribution of the numbers of $v$-structures.
To compare Algorithm~\ref{mc} with Algorithm~\ref{acc}, we re-estimate
these four distributions for completed PDAGs in ${\cal S}_{100}^{150}$
via Algorithm~\ref{mc}.

For each distribution, in Figure~\ref{differnce}, we report the
estimates obtained by Algorithm~\ref{mc} with lines and the estimates
obtained by Algorithm~\ref{acc} with points in the main windows. The
differences of two estimates are shown in the sub-windows. The top
panel of Figure~\ref{differnce} displays the cumulative distributions
of proportions of directed edges. The second panel of this figure
displays the distributions of the numbers of chain components. The
third panel displays the distributions of maximum size of chain
components. The bottom panel displays the distribution of the numbers
of $v$-structures. We can see that the differences of three pairs of
estimates are small.

\begin{figure}

\includegraphics{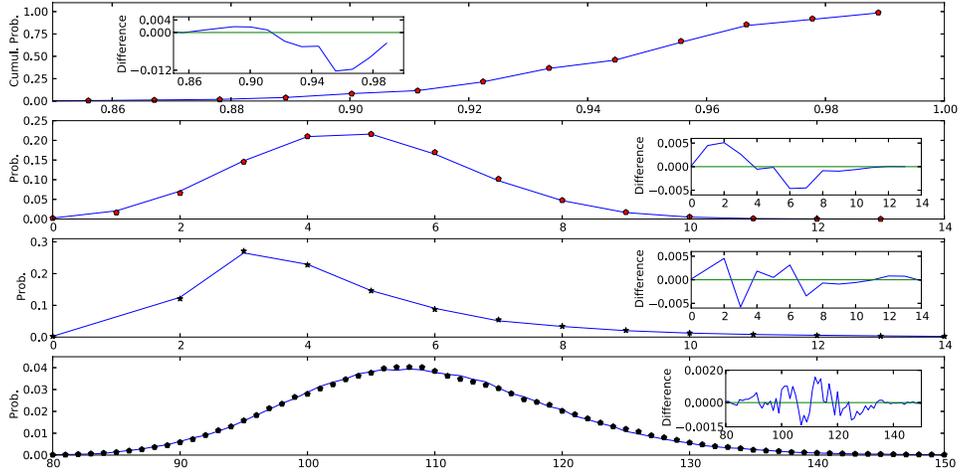}

\caption{Distributions for completed PDAGs in ${\cal S}_{100}^{150}$
estimated via Algorithm \protect\ref{mc} (plotted in lines) and the
accelerated version---Algorithm \protect\ref{acc} (plotted in points)
are shown in the main windows. The differences are shown in sub-windows.
Four panels (from top to bottom) display distributions of directed
edges, number of chain components, maximum size of chain components and
$v$-structures, respectively.}
\label{differnce}
\end{figure}

The average times used to generate a state of the Markov chain of
completed PDAGs in ${\cal S}_p^{1.5p}$ are shown in
Table~\ref{time}, in which $\alpha$ is the acceleration parameter used
in Algorithm~\ref{acc}. If $\alpha=1$, the Markov chain is generated
via Algorithm~\ref{mc}. The results suggest that the accelerated
version can speed up the approach nearly $\frac{1}{\alpha}$ times when $p=100$.

\subsection{Asymptotic properties of proposed estimators}
\label{exmixing}

We further illustrate the asymptotic properties of proposed estimators
of sparse completed PDAGs via simulation studies. We consider
${\mathcal S}_p^{1.5p}$ for $p=100,200$, $500$ and $1000$, respectively.
Let $f(u)$ be a discrete function of Markov equivalence class $u$,
where $u$ is a random variable distributed uniformly in ${\mathcal
S}_p^{1.5p}$. Let ${\mathbb{E}}(f)$ be the expectation of $f(u)$, and
we have
\[
{\mathbb{E}}(f)=\sum_{i} i{\mathbb P}(f=i).
\]
Proposition~\ref{estimator} shows that the estimator $\hat{\mathbb
P}(f=i)$ in equation \eqref{esteg} converges to ${\mathbb P}(f=i )$
with probability one.
We also have that the estimator defined as
\[
\hat{\mathbb{E}}(f)=\sum_{i}i\hat{\mathbb P}(f=i)=
\frac{\sum_{i} \sum_{t=1}^N iI_{\{f(e_t)=i\}}{M_t^{-1}}}{\sum_{t=1}^N
{M_t^{-1}}}=\frac
{\sum_{t=1}^N f(e_t){M_t^{-1}}}{\sum_{t=1}^N {M_t^{-1}}}
\]
converges to ${\mathbb{E}}(f)$ with probability one, where $\{
{e_t},M_t\}_{t=1,\ldots,N}$ is a Markov chain from Algorithm~\ref{mc}.

\begin{table}[b]
\tablewidth=242pt
\caption{The average time used to generate a completed PDAG in ${\cal
S}_p^{1.5p}$,
where $p$ is the number of vertices, $\alpha$ is the acceleration parameter,
$\kappa$ is the average time (seconds)}
\label{time}
\begin{tabular*}{\tablewidth}{@{\extracolsep{\fill}}lccccc@{}}
\hline
$p$& 100&100 & 200 & 500& 1000 \\
$\alpha$& 1& 0.1 &0.05 & 0.01 & 0.001 \\
$\kappa$ (seconds)& 0.22 &0.032 &0.113 &0.28 &0.72 \\
\hline
\end{tabular*}
\end{table}

\begin{figure}

\includegraphics{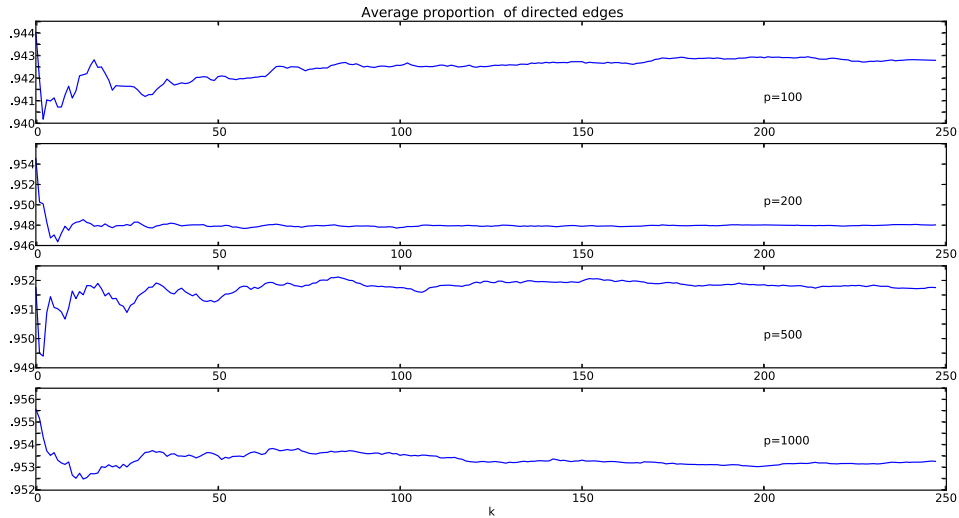}

\caption{Four sequences of average proportions of directed edges in
completed PDAGs in ${\cal S}_p^{1.5p}$ with $p=100,200,500$ and $1000$,
estimated via Algorithm \protect\ref{acc} and the first $5000k$ steps
of the Markov chains, where k is shown in x-axis.}
\label{mixingdirected}
\end{figure}

We generate some sequences of Markov equivalence classes $\{{e_t},\hat
{ M}_t\}$ with length of $N=1.25\times10^6$ via Algorithm~\ref{acc}
and divide each sequence into 250 blocks. Set $f(u)$ to be the
proportion of directed edges in $u$, we estimate ${\mathbb{E}}(f)$
using cumulative data in the first $k$ blocks as
\[
\hat{\mathbb{E}}(f)_k = \Biggl( \sum_{t = 1}^{k \times j }
{f(e_t)}\hat{M}_t^{-1} \Biggr)\bigg/\sum
_{t = 1}^{k \times j } \hat{M}_t^{-1},
\]
where $j=5\times10^3$. The simulation results are shown in Figure \ref
{mixingdirected}.
We can see that the estimates of proportions of directed edges
converge quickly
as $k$ increases.

%





\section{Conclusions and discussions}
\label{discussion}

In this paper, we proposed a reversible irreducible Markov chain on
Markov equivalence classes that can be used to study various properties
of a given set of interesting Markov equivalence classes.
Our experiments on Markov equivalence classes with sparse constraints
reveal useful information. For example, we find
that proportions of undirected edges and chain components in sparse completed
PDAGs are small even for Markov equivalence classes with thousands of vertices.


When some ``important'' but very rare equivalence classes are of interest,
it will be very hard to sample them in the proposed Markov chain. In this
case, we can constrain the space appropriately so that these Markov
equivalence classes are easy to be sampled. For example, it is nearly
impossible to sample equivalence classes with 300 vertices and 1 edge from
${\cal S}_{300}$. Fortunately, if we set the space to be ${\cal
S}_{300}^2$, sampling graphs with 1 edge is not difficult.


The sizes of Markov equivalence classes are the property most widely
discussed in the literature.
Due to space constraints, we have omitted several details in this paper
about determining the size of Markov equivalence classes and
calculating further properties of edges and vertices. We will discuss
these issues in a follow-up paper.
The proposed methods can potentially be extended to study other sets of
completed PDAGs besides ${\cal S}_n^p$. Some interesting sets include
(1) the completed PDAGs in which each vertex has at most
$d$ adjacent edges; (2) completed PDAGs in which each pair of vertices
is connected by a path along edges in the graph.

\begin{appendix}\label{app}
\section*{\texorpdfstring{Appendix: Preliminary results and proof of Theorem~\lowercase{\protect\ref{perfect}}}
{Appendix: Preliminary results and proof of Theorem 1}}

In this Appendix, we provide two preliminary results introduced by
Andersson~\cite{andersson1997characterization} and Chickering
\cite{chickering1995transformational,chickering2002learning},
respectively, in Appendix~\ref{appA.1}. These results are necessary to implement
our proposed approach technically and will be used in the proof of
Theorem~\ref{perfect}. Then we provide a proof of the main result of
this paper (Theorem~\ref{perfect}) in Appendix~\ref{appA.2}.

\subsection{Two preliminary results}\label{appA.1}

Some definitions and notation are introduced first. A graph is called a
\emph{chain graph} if it contains no partially directed cycles \cite
{lauritzen2002chain}.
A \emph{chord} of a cycle is an edge that joins two nonadjacent
vertices in the cycle.
An undirected graph is \emph{chordal}
if every cycle of length greater than or equal to $4$ possesses a
chord. A directed edge of a DAG is \emph{compelled} if it occurs in the
corresponding completed PDAG, otherwise, the directed edge is \emph
{reversible}, and the corresponding parents are reversible parents.
Recall $N_x$ be the set of all neighbors of $x$, $\Pi_x$ is the set of
all parent of $x$, $N_{xy}=N_x\cap N_y$ and $\Omega_{x,y}=\Pi_x \cap
N_y$ and the concept of ``strongly protected'' is presented in
Definition~\ref{spro}.

Lemma~\ref{essential} characterizes completed PDAGs that are used to
represent Markov equivalence classes \cite
{andersson1997characterization} and will be used in the proofs in Appendix~\ref{appA.2}.

\begin{lemma}[(Andersson~\cite{andersson1997characterization})]\label
{essential} A graph ${\cal C}$ is a completed PDAG of a directed
acyclic graph ${\cal D}$ if and only if ${\cal C}$ satisfies the following
properties:
\begin{longlist}[(iii)]
\item[(i)] ${\cal C}$ is a chain graph;
\item[(ii)] let $ {\cal C}_{\tau}$ be the subgraph induced by $\tau$. $
{\cal C}_{\tau}$ is chordal for every chain component $\tau$;
\item[(iii)] $w\rightarrow u -v$ does not
occur as an induced subgraph of ${\cal C}$;
\item[(iv)] every arrow $v\to u$ in ${\cal C}$ is strongly protected.
\end{longlist}
\end{lemma}

Lemma~\ref{chickervalid} shows the equivalent validity conditions for
\textbf{iu}$_2$, \textbf{du}$_1$, \textbf{id}$_2$, \textbf{dd}$_1$ and
\textbf{mv}$_1$ used in Definition~\ref{constrainedset}.

\begin{lemma}[(Validity conditions of some operators \cite
{chickering2002learning})] The necessary and sufficient validity
conditions of the operators with type of InsertU, DeleteU, InsertD,
DeleteD or MakeV are as follows:
\label{chickervalid}
\begin{itemize}
\item(InsertU) Let $x$ and $y$ be two vertices that are not adjacent
in $\cal C$. The operator InsertU $x-y$ is valid (equivalently, \textbf
{iu}$_2$ holds) if and only if $(\mathrm{iu}_{2.1})$ $\Pi_x = \Pi_y$,
$(\mathrm{iu}_{2.2})$ every undirected path from $x$ to $y$ contains a
vertex in $N_{xy}$.
\item(DeleteU) Let $x-y$ be an undirected edge in completed PDAG $\cal
C$. The operator DeleteU $x-y$ is valid (equivalently, \textbf{du}$_1$
holds) if and only if $(\mathrm{du}_{1.1})$ $N_{xy}$ is a clique in
${\cal C} $.
\item(InsertD) Let $x$ and $y$ be two vertices that are not adjacent
in $\cal C$. The operator InsertD $x\to y$ is valid (equivalently,
\textbf{id}$_2$ holds) if and only if $(\mathrm{id}_{2.1})$ $\Pi_x\neq\Pi
_y$, $(\mathrm{id}_{2.2})$ $\Omega_{x,y}$ is a clique,
$(\mathrm{id}_{2.3})$ every partially directed path from $y$ to $x$
contains at least one vertex in $\Omega_{x,y}$.
\item(DeleteD) Let $x\to y$ be a directed edge in completed PDAG $\cal
C$. The operator DeleteD of $x\to y$ is valid (equivalently, \textbf
{dd}$_1$ holds) if and only if $(\mathrm{dd}_{1.1})$ $N_y$ is a clique.
\item(MakeV) Let $x- z- y$ be any length-two undirected path in ${\cal
C}$ such that $x$ and $y$ are not adjacent. The operator MakeV $x\to
z\leftarrow y$ is valid (equivalently, \textbf{mv}$_1$ holds) if and
only if $(\mathrm{mv}_{1.1})$ every undirected path between $x$ and $y$
contains a vertex in $N_{xy}$.
\end{itemize}
\end{lemma}

\subsection{\texorpdfstring{Proof of Theorem \protect\ref{perfect}}{Proof of Theorem 1}}
\label{proof1}\label{appA.2}

Let ${\cal O}$ be the operator set defined in equation \eqref
{perfectset}; to prove Theorem~\ref{perfect}, which shows $\cal O$ is a
perfect operator set, we need to show $\cal O$ satisfies four
properties: validity, distinguishability, irreducibility and
reversibility. Equivalently, we just need to prove Theorem \ref
{valid}--\ref{irr} as follows:

\begin{theorem}\label{valid}
The operator set ${\cal O}$ is valid.
\end{theorem}
%
\begin{theorem}\label{disting}
The operator set ${\cal O}$ is distinguishable.
\end{theorem}
%
\begin{theorem}\label{reverseth}
The operator set ${\cal O}$ is reversible.
\end{theorem}
%
\begin{theorem}\label{irr}
The operator set ${\cal O}$
is irreducible.
\end{theorem}

Of the above four theorems, the most important and difficult is to
prove Theorem~\ref{reverseth}. We now show the proofs one by one.

\begin{pf*}{Proof of Theorem~\ref{valid}}
According to the definition of validity in Definition~\ref{validdef}
and the definition of ${\cal O}_{\cal C}$ in equation \eqref
{constraintedC}, all operators in $\mathit{InsertU}_{\cal C}$, $\mathit{DeleteU}_{\cal
C}$, $\mathit{InsertD}_{\cal C}$, $\mathit{DeleteD}_{\cal C}$ and $\mathit{MakeV}_{\cal C}$ are
valid. We just need to prove Lemma~\ref{remov}, which shows all
operators in $\mathit{RemoveV}_{\cal C}$ are valid.
\end{pf*}

\begin{lemma}\label{remov}
Let $x\to z \leftarrow y$ be a $v$-structure in completed PDAG ${\cal
C}$. If \emph{(\textbf{rv}$_1$)} $\Pi_{x}=\Pi_{y}$, \emph{(\textbf
{{rv}}$_2$)} $\Pi_{x}\cup N_{xy}=\Pi_{z}\setminus\{x,y\}$, and \emph
{(\textbf{{rv}}$_3$)} every undirected path between $x$ and $y$
contains a vertex in $N_{xy}$ hold, then the operator RemoveV $x\to z
\leftarrow y$ is valid and results in a completed PDAG in ${\cal
S}_p^n$ defined in equation \eqref{edgec}.
\end{lemma}

To prove Lemma~\ref{remov}, we will use Lemma~\ref{lemma32} given by
Chickering (Lemma 32 in~\cite{chickering2002learning}).

\begin{lemma}\label{lemma32}
Let ${\cal C}$ be any completed PDAG, and let $x$ and $y$ be any pair
of vertices that are not adjacent. Every undirected path between $x$
and $y$ passes through a vertex in $N_{xy}$ if and only if there exists
a consistent extension in which (1) $x$ has no reversible parents, (2)
all vertices in $N_{xy}$ are parents of $y$ and (3) $y$ has no other
reversible parents.
\end{lemma}

We now give a proof of Lemma~\ref{remov}.
\begin{pf*}{Proof of Lemma~\ref{remov}}
From Lemma~\ref{lemma32} and condition \textbf{rv}$_3$ in Lemma \ref
{remov}, there exists a consistent extension of $\cal C$, denoted by
$\cal D$, in which $x$ has no reversible parents, and the reversible
parents of $y$ are the vertices in $N_{xy}$. Because $y\to z$ occurs in
the completed PDAG, ${\cal C}$, $N_z$ and $N_y$ occur in different
chain components. We can orient the undirected edges adjacent to $z$
out of $z$. Then all vertices in $N_{z}$ are children of $z$ in $\cal D$.
Let ${\cal D}'$ be the graph obtained by reversing $y\to z$ in $
\cal D$ and ${\cal P}'$ be the PDAG obtained by applying the RemoveV
operator to ${\cal C}$. We will show that ${\cal D}'$ is a consistent
extension of ${\cal P}'$.

Clearly, ${\cal D}'$ and ${\cal P}'$ have the same skeleton.

We have that any $v$-structure that occurs in ${\cal D}$ but not in
${\cal P}'$ must include either the edge $x\to z$ or $y \to z$. Since
$\cal D$ is a consistent extension of ${\cal C}$, we have that all
$v$-structures in ${\cal D}$ are also in ${\cal C}$. From condition
\textbf{rv}$_2$, all parents of z other than $x$ and $y$ are adjacent
to $x$ and $y$. Hence $x\to z\leftarrow y$ is the only $v$-structure that
is directed into $z$ in ${\cal C}$. We have that all $v$-structures of
${\cal P}'$ are also in $\cal D$, and there is only one $v$-structure
$x\to z\leftarrow y$ that is in $\cal D$ but not ${\cal P}'$.

Since $y\to z$ is the unique edge that differs between $\cal D$ and
${\cal D}'$, we have that any $v$-structure that exists in ${\cal D}$ but
not in ${\cal D}'$ must include the edge $y \to z$, and any $v$-structure
that exists in ${\cal D}'$ but not in ${\cal D}$ must include the edge
$z\to y$. We have shown that $x\to z\leftarrow y $ is the only
$v$-structure in ${\cal D}$ that is directed into $z$. From the
construction of ${\cal D}$, we have that all compelled parents of $y$
in ${
\cal D}'$ are also parents of $z$, and all other parents are in
$N_{xy}$; from \textbf{rv}$_2$, they also are parents of $z$. There is
no $v$-structure that includes edge $z\to y$ in ${\cal D}'$. Hence, all
$v$-structures of ${\cal D}'$ are also in $\cal D$, and there is only one
$v$-structure $x\to z\leftarrow y$ that is in $\cal D$ but not ${\cal D}'$.

Hence, ${\cal D}'$ and ${\cal P}'$ have the same $v$-structures. It
remains to be shown that ${\cal D}'$ is acyclic.\vadjust{\goodbreak}

If ${\cal D}'$ contains a cycle, the cycle must contain the edges $z\to
y$ because $\cal D$ is acyclic. This implies there is a directed path
from $y$ to $z$ in $\cal D$. By construction, all vertices in $N_z$ are
children of $z$ in ${\cal D}'$. So, this path must include a compelled
parent of $z$; denote it by $u$. If $u\neq x$, from condition \textbf
{rv}$_2$, $u\in\Pi_y \cup N_{xy}$; by the construction of $\cal D$,
we have $u\in\Pi_y$. Thus, there is no path from $y$ to $z$ that
contains~$u$. If $u= x$, by construction, the path must contain a
compelled parent $v$ of $x$. From condition \textbf{rv}$_1$, $v\in\Pi
_y$. Thus, there is no path from $y$ to $z$ contains $v$. We get that
${\cal D}'$ is acyclic. Thus ${\cal D}'$ is a consistent extension of
${\cal P}'$ and the operator RemoveV $x\to z \leftarrow y$ is valid.
\end{pf*}

\begin{pf*}{Proof of Theorem~\ref{disting}}
For any completed ${\cal C} \in{\cal S}_p^n$, we need to show that
different operators in ${O}_{\cal C}$ result in different completed
PDAGs. For any valid operator $o\in \mathit{InsertU}_{\cal C}$, say InsertU
$x-y$, denoted as $o$, the resulting completed PDAG of $o$ contains the
undirected edge $x-y$. We have that all other operators in ${O}_{\cal
C}$ except for InsertD $x\to y$ and Insert $x \leftarrow y$ (if they
are also valid) will result in completed PDAGs with skeletons different
than the resulting completed PDAG of~$o$. Thus, these operators cannot
result in the same completed PDAG as $o$. If InsertD $x\to y$ or Insert
$x \leftarrow y$ is valid, the resulting completed PDAGs of them
contain $x\to y$ or $x \leftarrow y$. These two resulting completed
PDAGs have at least a compelled edge different than the resulting
completed PDAG of $o$. Thus there is no operator in ${O}_{\cal C}$ that
can result in the same completed PDAG as $o$.

Similarly, we can show for any operator in ${\cal O}_{{\cal C}}$,
different operators will result in different completed PDAGs because
they will have distinct skeletons, compelled edges or $v$-structures.
\end{pf*}

\begin{pf*}{Proof of Theorem~\ref{reverseth}}
Let $\cal C$ be any completed PDAG in ${\cal S}_p^n$, $o\in{\cal
O}_{\cal C}$ be an operator on $\cal C$. The operator $o'
\in{\cal O}$ is the \emph{reversible operator} of $o$ if $o'$ can
transfer the resulting completed PDAG of $o$ back to $\cal C$. To prove
Theorem~\ref{reverseth}, we just need to show each operator in ${\cal
O}_{\cal C}$ defined in equation \eqref{perfectset} has a reversible
operator in~$\cal O$.
Equivalently, we prove Lemmas~\ref{re1},~\ref{re2},
\ref{re3},~\ref{re4},~\ref{re5} and~\ref{re6} to show the
reversibility for six types of operators, respectively.
\end{pf*}

\begin{lemma}\label{re1}
For any operator $o\in{\cal O}_{{\cal C}}$ denoted by ``InsertU
$x-y$,''
the operator ``DeleteU $x-y$'' is the reversible operator of $o$.
\end{lemma}

\begin{lemma}\label{re2}
For any operator $o\in{\cal O}_{{\cal C}}$ denoted by ``DeleteU
$x-y$,'' the operator ``InsertU $x-y$'' is the reversible operator of $o$.
\end{lemma}

\begin{lemma}\label{re3}
For any operator $o\in{\cal O}_{{\cal C}}$ denoted by ``InsertD $x\to
y$,'' the operator ``DeleteD $x\to y$'' is the reversible operator of $o$.
\end{lemma}

\begin{lemma}\label{re4}
For any operator $o\in{\cal O}_{{\cal C}}$ denoted by ``DeleteD $x\to
y$,'' the operator ``InsertD $x\to y$'' is the reversible operator of $o$.\vadjust{\goodbreak}
\end{lemma}

\begin{lemma}\label{re5}
For any operator $o\in{\cal O}_{{\cal C}}$ denoted by ``MakeV $x\to
z\leftarrow y$,'' the operator ``RemoveV $x\to z\leftarrow y$'' is the
reversible operator of $o$.
\end{lemma}

\begin{lemma}\label{re6}
For any operator $o\in{\cal O}_{{\cal C}}$ denoted by ``RemoveV $x\to
z\leftarrow y$,'' the operator ``MakeV $x\to z\leftarrow y$'' is the
reversible operator of $o$.
\end{lemma}

Before giving proofs of these six lemmas, We first provide several
results shown in Lemmas~\ref{cq},~\ref{re000},~\ref{redecs} and
\ref{re0str}.

\begin{lemma}\label{cq}
Let graph ${\cal C}$ be a completed PDAG, $\{w,v,u\}$ be three
vertices that are adjacent each other in ${\cal C}$. If there are two
undirected edges in $\{w,v,u\}$, then the third edge is also undirected.
\end{lemma}

\begin{pf}
If the third edge is directed, there is a directed cycle like $w-
v-u\to w$. From Lemma~\ref{essential}, we know that ${\cal C}$ is a
chain graph, so there is no directed circle in ${\cal C}$.
\end{pf}

\begin{lemma}\label{re000}
Let ${\cal C}_1$ be the resulting completed PDAG obtained by inserting
a new edge between $x$ and $y$ in $\cal C$. If there is at least one
edge $v \to u$ that is directed in ${\cal C}$ but not directed in
${\cal C}_1$, then there exists a vertex $h$ that is common child of
$x$ and $y$ such that $x\to h$ and $y\to h$ in ${\cal C}$ become
undirected in ${\cal C}_1$.
\end{lemma}

\begin{pf}
According to Lemma~\ref{essential}, an edge is directed in a completed
PDAG if and only if it is strongly protected. Thus, we have that at
least one case among (a), (b), (c), (d) in Figure~\ref{spro111} occurs
in ${\cal C}$ but not in ${\cal C}_1$ for $v\to u$. We will show that
either Lemma~\ref{re000} holds, or there exists a parent of $u$,
denoted as $u_1$, such that $u_2\to u_1$ occurs in ${\cal C}$ but not
in ${\cal C}_1$, where $u_2$ is a parent of $u_1$. We denote the latter
result as (*).

Suppose case (a) in Figure~\ref{spro111} occurs in ${\cal C} $ but not
in ${\cal C}_1$. Because $v\to u$ becomes undirected in ${\cal C}_1$,
we have that $w\to v$ must be undirected in ${\cal C}_1$ since $w$ and
$u$ are not adjacent. Set $u_1=v$ and $u_2=u$, and we have that (*) holds.

Suppose case (b) in Figure~\ref{spro111} occurs in ${\cal C}$ but not
in ${\cal C}_1$. If the pair $\{v,w\}$ is not $\{x,y\}$, $v\to
u\leftarrow w$ is a $v$-structure in ${\cal C}$. We have that $v\to u$
occurs in ${\cal C}_1$. This is a contradiction. If $\{v,w\}$ is $\{
x,y\}$, we have that Lemma~\ref{re000} holds ($h=u$).

Suppose case (c) in Figure~\ref{spro111} occurs in ${\cal C}$ but not
in ${\cal C}_1$. Either $v\to w$ or $w\to u$ occurs in ${\cal C}$ but
not in ${\cal C}_1$. If it is $v\to w$, by setting $u_2=v$ and $u_1=w$,
we have (*) holds. If it is $w\to u$, both $v-u$ and $w-u$ in ${\cal
C}_1$, so $x-u$ also must be in ${\cal C}_1$. We also have that (*) holds.

Suppose case (d) in Figure~\ref{spro111} occurs in ${\cal C}$ but not
in ${\cal C}_1$. If the pair $\{w,w_1\}$ is $\{x,y\}$, Lemma \ref
{re000} holds ($h=u$). Otherwise, $w\to u\leftarrow w_1$ must occur in
both ${\cal C}_1$ and ${\cal C}$ and the edge $v\to u$ is still
strongly protected in ${\cal C}_1$, yielding a contradiction.

If (*) holds, we have that there is a directed path $u_2\to u_1\to u$
such that $u_2\to u_1$ occurs in ${\cal C}$ but not ${\cal C}_1$.
Iterating, we can get a directed path $u_k\to u_{k-1}\cdots\to u$ of
length $k-1$ without undirected edges such that $u_{k}\to u_{k-1}$
occurs in ${\cal C}$ but not in ${\cal C}_1$ if Lemma~\ref{re000} does
not hold in each step. Because ${\cal C}$ is a chain graph without
directed circle, the procedure will stop in finite steps and Lemma \ref
{re000} will hold eventually.
\end{pf}

From the proof of Lemma~\ref{re000}, we have that $u$ should be a
descendant of $x$ and~$y$, so we can get the following Lemma~\ref{redecs}.

\begin{lemma}\label{redecs}
Let ${\cal C}$ be any completed PDAG, and let $\cal P$ denote the PDAG
that results from adding a new edge between $x$ and $y$. For any edge
$v\to u$ in ${\cal C}$ that does not occur in the resulting completed
PDAG extended from $\cal P$, there is a directed path of length zero or
more from both $x$ and $y$ to $u$ in ${\cal C}$.
\end{lemma}

\begin{lemma}\label{re0str}
Let $ \mathit{InsertU}_{\cal C}$ and $\mathit{DeleteU}_{\cal C}$ be the operator
sets defined in Definition~\ref{constrainedset}, respectively. For any
$o$ in $\mathit{InsertU}_{\cal C}$ or in $\mathit{DeleteU}_{\cal C}$, where ${\cal
P}'$ is the modified graph of $o$ that is obtained by applying $o$ to
${\cal C}$, we have that ${\cal P}'$ is a completed PDAG.
\end{lemma}

\begin{pf}
We just need to check whether ${\cal P}'$ satisfies the four conditions
in Lemma~\ref{essential}.

(i):
For any $o\in\mathit{DeleteU}_{\cal C}$, denoted as DeleteD $x-y$, let ${\cal
P}'$ be the modified graph obtained by deleting $x-y$ from ${\cal C}$.

If there is a directed cycle in ${\cal P}'$, it must be a directed
cycle in ${\cal C}$, which is a contradiction. Thus there is no
directed cycle in ${\cal P}'$, and ${\cal P}'$ is a chain graph.

If there exists an undirected cycle of length greater than 3 without a
chord in~${\cal P}'$, the cycle must contain both $x$ and $y$;
otherwise, this cycle occurs in ${\cal C}$. If the length of the cycle
is 4, the other two vertices are in $N_{xy}$; we have that the cycle
has a chord since $N_{xy}$ is a clique in ${\cal C}$. If the cycle in
${\cal P}'$ has length greater than 4 without a chord, we have that
$x-y$ is the unique chord of this cycle in ${\cal C}$. However, this
would imply that there is a cycle of length greater than 3 without a
chord in~${\cal C}$, a~contradiction. Thus, there is no undirected
cycle with length greater than 3 in ${\cal P}'$, so every chain
component of ${\cal P}'$ is chordal.

Suppose that $\cdot\to\cdot-\cdot$ occurs as an induced subgraph
of ${\cal P}'$; it must be $x \to\cdot-y$ (or $y \to\cdot-x$).
However, in this case, $x\to\cdot-y-x$ (or $y \to\cdot-x-y$) would be
a directed cycle in ${\cal C}$. Thus the induced subgraph like $\cdot
\to\cdot-\cdot$ does not occur as an induced subgraph of ${\cal P}'$.

Finally, all directed edges in ${\cal P}'$ will be strongly protected;
by the definition of strong protection, all directed edges in $\cal C$
will remain strongly protected when an undirected edge is removed.\vadjust{\goodbreak}

(ii): For any $o\in\mathit{InsertU}_{\cal C}$, denoted as InsertU $x-y$,
${\cal P}'$ is the modified graph of $o$.

If there is a directed cycle in ${\cal P}'$, it must contain $x-y$;
otherwise this cycle is also in ${\cal C}$. We can suppose that there
exists a partially directed path from $x$ to $y$ in~${\cal C}$. Denote
the adjacent vertex of $y$ in the path as $u$. Let $u$ be the vertex
adjacent to $y$ in the path. We have $u\notin\Pi_y$; otherwise, from
the condition $\Pi_x=\Pi_y$ in Lemma~\ref{chickervalid}, $u$~would also
be in $\Pi_x$, so there would be a partially directed cycle from $x$ to
$x$ in ${\cal C}$. Hence the directed path must have the form $x\cdots
\to\cdots u-y$. This would induce a subgraph like $a\to b-v$ in ${\cal
C}$, a contradiction. Consequently, ${\cal P}'$ is a chain graph.

If there exists an undirected cycle of length greater than 3 without a
chord in~${\cal P}'$, the cycle must contain $x$ and $y$, and there
must be an undirected path from $x$ to $y$ in~${\cal C}$; otherwise,
the cycle would also be in ${\cal C}$. From Lemma~\ref{chickervalid},
every undirected path from $x$ to $y$ contains a vertex in $N_{xy}$, so
every undirected path of length greater than two has a chord. Thus,
every undirected path of length greater than 3 from $x$ to $y$ in
${\cal P}'$ has a chord. This implies that every chain component of
${\cal P}'$ is chordal.

Suppose that a subgraph like $\cdot\to\cdot-\cdot$ occurs as an
induced subgraph of ${\cal P}'$. Since $\Pi_x= \Pi_y$ in $\cal C$, the
induced subgraph is not $\cdot\to x-y$ (or $\cdot\to y-x$). Thus, the
induced subgraph like $\cdot\to\cdot-\cdot$ also occurs in ${\cal C}$.
This is a contradiction since $\cal C$ is a completed PDAG, yielding a
contradiction.

From Lemma~\ref{re000} and the condition \textbf{iu}$_3$ in Definition
\ref{constrainedset}, all directed edges in ${\cal C}$ are also
directed in ${\cal C}_1$. This implies that all directed edges in
${\cal P}$ are still compelled, and are thus strongly protected.
\end{pf}

We now give proofs of Lemmas~\ref{re1},~\ref{re2},
\ref{re3},~\ref{re4},~\ref{re5} and~\ref{re6}, one by one.

\begin{pf*}{Proof of Lemma~\ref{re1}}
Because the operator ``InsertU $x-y\mbox{''} = o\in{\cal O}_{{\cal C}}$ is
valid and ${\cal C}_1$ is the resulting completed PDAG of $o$, we have
that $x-y$ occurs in ${\cal C}_1$. We just need to show that the common
neighbors of $x$ and $y$, denoted as $N_{xy}$, form a clique in ${\cal C}_1$.

If $N_{xy}$ is empty set or has only one vertex, the condition that
$N_{xy}$ is a clique in ${\cal C}_1$ holds.

If there are two different vertices $z,u\in N_{xy}$ in ${\cal C}_1$, we
have that $x-z-y$ and $x-u-y$ form a cycle of length of 4 in ${\cal
C}_1$. The cycle is also in ${\cal C}$. Since the edge $x-y$ does not
exist in ${\cal C}$ and ${\cal C}$ is a completed PDAG in which all
undirected subgraphs are chordal graphs, we have that $z-u$ occurs in
${\cal C}$, so $z$ and $u$ are adjacent in ${\cal C}_1$. Hence the
condition that $N_{xy}$ is a clique in ${\cal C}_1$ holds.
\end{pf*}

\begin{pf*}{Proof of Lemma~\ref{re2}}
We need to show the operator $o'$:= InsertU $x-y$ satisfies the
conditions \textbf{iu}$_1$, \textbf{iu}$_2$ and \textbf{iu}$_3$ in
Definition~\ref{constrainedset} for completed PDAG ${\cal C}_1$ and
that the resulting completed PDAG of $o'$ is ${\cal C}$.

The condition \textbf{iu}$_1$ clearly holds, since $x-y$ exists in
${\cal C}_1$ but not in ${\cal C}$.
Lemma~\ref{re0str} implies that the graph obtained by deleting $x-y$
from $\cal C$ is the completed PDAG~${\cal C}_1$. Thus, the graph
obtained by inserting $x-y$ into ${\cal C}_1$ is~$\cal C$. This implies
that InsertU $x-y$ is valid, and the condition \textbf{iu}$_2$ holds.

Lemma~\ref{re0str} implies that the condition iu$_3$ also holds.
\end{pf*}

\begin{pf*}{Proof of Lemma~\ref{re3}}
I will first show that there is no undirected edge $y-w$ that occurs
in both ${\cal C}$ and ${\cal C}_1$.
If $w-y$ occurs in ${\cal C}$, since $x$ and $y$ are not adjacent in
${\cal C}$, $x\to w-y$ does not occur in ${\cal C}$. There are three
possible configurations between $x$ and $w$ in ${\cal C}$: (1) $x$ is
not adjacent to $w$, (2) $w\to x$ and (3) $x-w$. If $x$ is not adjacent
to $w$ in ${\cal C}$, inserting $x\to y$ will result in $y\to w$ in
${\cal C}_1$. If $w\to x$ is in~${\cal C}$, inserting $x\to y$ will
result in $w \to y$ in ${\cal C}_1$. If $x-w$ in ${\cal C}$, there is
an undirected path from $y$ to $x$; that is, the first condition for
InsertD to be valid, according to Lemma~\ref{chickervalid}, does not hold.
Thus we get that there is no undirected edge $y-w$ that occurs in both
${\cal C}$ and ${\cal C}_1$.

For any $w\in N_{y}$ in ${\cal C}_1$, the edge between $w$ and $y$ is
directed in ${\cal C}$; that is, either $w\to y$ or $y\to w$ occurs
in~$\cal C$. If $y\to w$ is in ${\cal C}$, there are three possible
configurations between $x$ and $w$ in ${\cal C}$: (1) $x$ is not
adjacent to $w$, (2) $w\to x$ and (3) $x\to w$. If $x$ and $w$ are not
adjacent in ${\cal C}$, inserting $x\to y$ will result in $y\to w$
in~${\cal C}_1$. If $w\to x$ occurs in ${\cal C}$, inserting $x\to y$
is not valid for ${\cal C}$ since there would be a directed path from
$y$ to $x$. If $x \to w$ occurs in ${\cal C}$, $w$ is common child of
$x$ and~$y$, so from condition id$_3$, $y\to w$ occurs in ${\cal C}_1$
and $w \notin N_y$ in ${\cal C}_1$. Thus, we have that $w\to y$ must be
in ${\cal C}$.

If there is another vertex $v\in N_{y}$ in ${\cal C}_1$, $v\to y$ must
also be in ${\cal C}$. If $v$ and $w$ are not adjacent, $v\to
y\leftarrow w$ forms a $v$-structure both in ${\cal C}$ and in ${\cal
C}_1$. $w\to y$ must occur in ${\cal C}_1$ and, consequently, $w \notin
N_y$ in ${\cal C}_1$ yielding a a contradiction. Thus, we know that any
two vertices in $N_y$ are adjacent in ${\cal C}$. $N_y$ is therefore a
clique in~${\cal C}_1$, and the operator DeleteD $x\to y$ is valid for
${\cal C}_1$; that is, the condition \textbf{id}$_1$ in Definition \ref
{constrainedset} holds.

Denote the modified PDAG of operator DeleteD $x\to y$ of ${\cal C}_1$
as ${\cal P}'$.
We need to show that the corresponding completed PDAG of ${\cal P}'$ is
$\cal C$. Equivalently, we just need to show ${\cal P}'$ and $\cal C$
have the same skeleton and $v$-structures. Clearly, ${\cal P}'$ and $\cal
C$ have the same skeleton. If there is a $v$-structure in $\cal C$, but
not in ${\cal C}_1$, it must be $x\to u\leftarrow y$, where $u$ is a
common child of $x$ and~$y$. From condition \textbf{id}$_3$ in
Definition~\ref{constrainedset}, $x\to u$ and $y\to u$ also occur
in~${\cal C}_1$, so, these $v$-structures also exist in~${\cal P}'$. This
implies that all $v$-structures of $\cal C$ are also in ${\cal P}'$.
Moreover, the $v$-structures in ${\cal C}_1$ but not in $\cal C$ must be
$x\to y\leftarrow v$, where $v$ is parent of $y$, and $x$ and $v$ are
not adjacent in ${\cal C}_1$. Clearly, after we delete $x\to y$ from
${\cal C}_1$, these $v$-structures will not exist in~${\cal P}'$. This
implies that all $v$-structures of ${\cal P}'$ are in $\cal C$. So,
${\cal P}'$ and $\cal C$ have the same $v$-structures.

For any $v\to y$ in ${\cal C}_1$, if $v- y$ is in ${\cal C}$, $v$ must
be parent of $x$. If $x$ and $v$ are not adjacent, inserting $x\to y$
to ${\cal C}$ will result in $y \to v$ in ${\cal C}_1$. Moreover,
$x-v-y$ does not exist in $\cal C$ since InsertD $x\to y$ is a valid
operator, and $x\to v- y$ does not occur in ${\cal C}$. Thus, for any
$v$ that is a parent of $y$ but not a parent of $x$, the directed edge
$v\to y$ also occurs in the resulting completed PDAG ${\cal C}$. That
is, the condition id$_2$ in Definition~\ref{constrainedset} holds.
\end{pf*}

\begin{pf*}{Proof of Lemma~\ref{re4}}
To prove this lemma, we first introduce Lemmas~\ref{shortestpath} and
\ref{fdel}.
Let $L=(u_1,u_2,\ldots,u_k)$ be a partially directed path
from $u_1$ to $u_k$ in a graph. A path $L_2=(u^1,\ldots,u^k)$ is a
sub-path of $L_1$ if all vertices in $L_1$ are in $L$ and have the same
order as in $L$. We say that a partially directed path is shortest if
it has no smaller sub-path.
\end{pf*}

\begin{lemma}\label{shortestpath}
Let ${\cal C}$ be a completed PDAG, and let $L_1$ be a partially
directed path from $y$ to $x$ in $\cal C$.
Then there exists a shortest sub-path of $L_1$, denoted as
$L_2=y-u_1-\cdots-u_k\to\cdots\to x$,
in which there exists a $k$ such that all edges occurring before $u_k$
in the path are undirected, and all edges occurring after $u_k$ are directed.
\end{lemma}

\begin{pf}
We just need to show that a directed edge must be followed by a
directed edge in the shortest sub-path. If not,
$u_i\to u_{i+1}-u_{i+2}$ occurs in $L_2$. Because ${\cal C}$ is a
completed PDAG, $u_i$ and $u_{i+2}$ must be adjacent; otherwise
$u_{i+1}\to u_{i+2}$ occurs in ${\cal C}$.
If $u_i \to u_{i+2}$ occurs in ${\cal C}$, $L_2$ is not a shortest
path. If $u_i \leftarrow u_{i+2}$ occurs in ${\cal C}$, $u_{i+1}
\leftarrow u_{i+2}$ must be in ${\cal C}$.
\end{pf}

\begin{lemma}\label{fdel}
If the graph ${\cal P}_1$ obtained by deleting $a\to b$ from a
completed PDAG ${\cal C}$ can be extended to a new completed PDAG,
${\cal C}_1$, then we have that for any directed edge $x\to y$ in
${\cal C}$, if $y$ is not $b$ or a descendent of $b$, then $x\to y$
occurs in~${\cal C}_1$.
\end{lemma}

\begin{pf}
Because $x\to y$ occurs in ${\cal C}$, so it is strongly protected in
${\cal C}$. If $x\to y$ does not occur in ${\cal C}_1$, it is not
strongly protected in ${\cal C}_1$ from Lemma~\ref{essential}. From the
definition of strongly protected, we know that the four cases in
Figure~\ref{spro111} in which $v\to u$ is strongly protected do not
involve any descendant of $u$. Thus, if $x\to y$ is not compelled in
${\cal C}_1$, there must exist a directed edge $w\to z$ between two
nondescendants of $y$ such that the edges between nondescendants of
$z$ are strongly protected, and $w-z$ is no longer strongly protected
in ${\cal P}_1$. Because ${\cal P}_1$ is obtained by deleting $a\to b$,
$z$ is nondescendant of $b$, we have that $w\to z$ is strongly
protected in ${\cal P}_1$, yielding a contraction.~%
\end{pf}
We now give a proof of Lemma~\ref{re4}:
\begin{pf*}{Proof of Lemma~\ref{re4}}
Since ${\cal C}\in{\cal S}_p^n$, we have $n_{{\cal C}_1}<n$. That is,
the condition id$_1$ in Definition~\ref{constrainedset} holds for
InsertD $x\to y$ of ${\cal C}_1$.

For any undirected edge $w-y$ in ${\cal C}$, $x$ must be parent of $w$;
otherwise the edge between $y$ and $w$ is directed. Then deleting\vadjust{\goodbreak} $x\to
y $ from ${\cal C}$ will result in $w\to y$ in~${\cal C}_1$. Thus, we
have that all $N_y$ in ${\cal C}$ become parents of $y$ in ${\cal C}_1$.
From the condition \textbf{dd}$_2$, the parents of $y$ but not $x$ in
${\cal C}$ are also parents of $y$ in~${\cal C}_1$.
If there is a partially directed path from $y$ to $x$ in ${\cal C}_1$,
then the vertex adjacent to $y$
in this path must be a child of $y$ or a vertex that is parent of $y$
and $x$ in ${\cal C}$. We will show that if the vertex is not a parent
of $y$ and $x$ in~${\cal C}$, there exists a contradiction.

If there is a partially directed path from $y$ to $x$ in ${\cal C}_1$,
we can find a shortest partially directed path like $y-u_1-\cdots-u_k\to
\cdots\to x$ from Lemma~\ref{shortestpath}, denoted as $L_1$. Any
directed edge, say $u_i\to u_{i+1}$, in $L_1$ does not become
$u_i\leftarrow u_{i+1}$ in~${\cal C}$. If $L_1$ does not include
undirected edges in ${\cal C}_1$, we have that the vertices of $L_1$
form a partially directed cycle in ${\cal C}$. We just need to show
that the vertices of the undirected path $L_1$ also form a partially
directed path in ${\cal C}$.

Suppose $y\to u_1$ occurs in ${\cal C}$. If $u_1-u_{2}$ is undirected
in ${\cal C}$, then $y\to u_{2}$ must occur in ${\cal C}$, and
consequently, $L_1$ will not be shortest in ${\cal C}_1$. If $u_2\to
u_{1}$ occurs in ${\cal C}$, there exists a $v$-structure $u_2\to
u_{1}\leftarrow y$ in ${\cal C}_1$; otherwise $u_2$ and $y$ are
adjacent, and $L_1$ is not the shortest path in ${\cal C}_1$. Thus,
$u_1\to u_2$ must occur in $\cal C$.
In this manner, we get that all edges in $y-u_1-\cdots-u_k\to\cdots\to
x$ are directed in ${\cal C}$ and are directed
from $u_i\to u_{i+1}$. This implies that there exists a partially
directed cycle in~${\cal C}$. So, $u_1$ must be a parent of $y$ and
$x$ in $\cal C$.
We have $u_1\in\Omega_{xy}$ and every partially directed path of
${\cal C}_1$ from $y$ to $x$ contains at least one vertex in $\Omega_{xy}$.

Since all vertices in $\Omega_{xy}$ in ${\cal C}_1$ are parents of $x$
and $y$ in $\cal C$, if there are two vertices, say $w_1,w_2 \in\Omega
_{xy}$, that are not adjacent, the subgraph $w_1\to y\leftarrow w_2$
could be a $v$-structure in ${\cal C}_1$. So, all vertices in $\Omega
_{xy}$ in ${\cal C}_1$ are adjacent and $\Omega_{xy}$ is a clique.

We have that the parents of $y$ in ${\cal C}_1$ ($(\Pi_y)_{{\cal
C}_1}$) are in the union of the parents and neighbors of $y$ in ${\cal
C}$ ($(\Pi_y\cup N_y)_{{\cal C}_1}$).
If there is at least one neighbor $u$ of $y$ in ${\cal C}$, $u$ must be
child of $x$ in ${\cal C}$ and
parent of $y$ in ${\cal C}_1$, so parents of $x$ and $y$ are not the same.
If there is no neighbor of $y$ in ${\cal C}$,
the parents of $y$ in ${\cal C}_1$ are the same as in~${\cal C}$,
except those vertices that are parents of $x$, that is, $(\Pi_y-\Pi
_x)_{{\cal C}_1}=(\Pi_y-\Pi_x)_{{\cal C}}$. At the same time, from
Lemma~\ref{fdel},
the parents of $x$ in ${\cal C}_1$ are also the parents of $x$ in~$\cal
C$. Thus, the parents of $x$ and $y$ are not the same in ${\cal C}_1$.
From Lemma~\ref{chickervalid}, we have that InsertD $x\to y$ is valid
for~${\cal C}_1$, and condition \textbf{id}$_2$ holds.

Denote the modified PDAG of operator InsertD $x\to y$ of ${\cal C}_1$
as ${\cal P}'$. We need to show that the corresponding completed PDAG
of ${\cal P}'$ is $\cal C$. Equivalently, we just need to show that
${\cal P}'$ and $\cal C$ have the same skeleton and $v$-structures.
Clearly, ${\cal P}'$~and $\cal C$ have the same skeleton. A
$v$-structure that is in $\cal C$ but not in ${\cal C}_1$ must have the
form $x\to y \leftarrow u$, where $u$ is parent of $y$ but not adjacent
to $x$. From condition \textbf{dd}$_2$ in Definition
\ref{constrainedset}, $u\to y$ also occurs in ${\cal C}_1$, so such a
$v$-structure must also exist in~${\cal P}'$. This implies that all
$v$-structures of $\cal C$ are also in ${\cal P}'$. Moreover, the
$v$-structures in ${\cal C}_1$ but not in $\cal C$ must have the form
$x\to v\leftarrow y$, where $v$ is a common child of $y$ and $x$ in
${\cal C}_1$. Clearly, after we insert $x\to y$ to ${\cal C}_1$, this
is no longer a $v$-structure in ${\cal P}'$ implying that all
$v$-structures of ${\cal P}'$ are in $\cal C$. Thus, ${\cal P}'$ and
$\cal C$ have the same $v$-structures.

Let the modified graph of DeleteD $x\to y$ from $\cal C$ be $\cal P$;
we know that $\cal P$ and ${\cal C}_1$ have the same $v$-structures.
Thus, for any $u$ that is a common child of $x$ and $y $ in ${\cal
C}_1$, $x\to u\leftarrow y$ is a $v$-structure in $\cal P$. This implies
that $y\to u$ occurs in $\cal C$ and the condition \textbf{id}$_3$ hold.
\end{pf*}

\begin{pf*}{Proof of Lemma~\ref{re5}}
Since $x,z$ and $y$ are in the same chain component of ${\cal C}$, they
have the same parent set in ${\cal C}$. The modified graph of $o'$ has
the same skeleton and $v$-structures as ${\cal C}_1$ because all
compelled edges in ${\cal C}$ remain compelled in ${\cal C}_1$. We just
need to prove that the operator $o'$ is valid and equivalently to prove
that the conditions \textbf{rm}$_1 $, \textbf{rm}$_2 $ and \textbf
{rm}$_3$ hold for ${\cal C}_1$.

We now show that the condition \textbf{rm}$_1$, $x$ and $y$ have the
same parents in
${\cal C}_1$ holds. Because $x$ and $y$ have the same parents in $\cal
C$, and all directed edges in $\cal C$ occur in~${\cal C}_1$, we just
need to consider the neighbors of $x$ or $y$.
Let $w-y$ be any undirected edge in ${\cal C}$, we consider the edges
between $w$ and $x$ or $z$:
\begin{longlist}[(4)]
\item[(1)] If both $w-z$ and $x-w$ occur in ${\cal C}$, $w-y$ and $w-x$
must be undirected in
${\cal C}_1$.
\item[(2)] If $w-z$ occurs but $x-w$ does not occur in ${\cal C}$,
$z\to w $ and $y\to w$ must be in ${\cal C}_1$.
\item[(3)] If $x-w$ occurs but $w-z$ does not occur in ${\cal C}$,
there is an undirected cycle of length 4 without a chord in
${\cal C}$. Thus, this case will not occur.
\item[(4)] If neither $w-z$ nor $x-w$ occur in ${\cal C}$, and there is
no undirected path other than $w-y-z$ from $w$ to $z$
in ${\cal C}$, then $w-y$ occurs in ${\cal C}_1$. If there exists
another undirected path from $w$ to $z$, there must exist an undirected
path of length 2 like $w-u'-z$ in ${\cal C}$,
and $y$ is adjacent to $u'$. In this case, $y-w$ occurs in ${\cal C}_1$
when $x-u'$ occurs and $y\to w$ occurs when $x$, and $u'$ are not adjacent.
\end{longlist}

Thus, there are no neighbors of $y$ in $\cal C$ that become parents of
$y$ in ${\cal C}_1$; that is, $y$ has the same parents in both ${\cal
C}_1$ and ${\cal C}$.
Similarly, $x$ has the same parents in both ${\cal C}_1$ and ${\cal
C}$. we get $x$ and $y$ have the same parents in
${\cal C}_1$, and the condition \textbf{rm}$_1$ holds.

All parents of $x$ must also be parents of $z$ in ${\cal C}_1$ since
they are in the same chain component.
For any $w\in N_{xy}$, $w-z$ also occurs in ${\cal C}$; otherwise
$x-z-y-w-x$ would form cycle of length 4 without a chord.
We have $w\to z$ must be in ${\cal C}_1$, otherwise a new $v$-structure
will occur in ${\cal C}_1$. Thus, we have $\Pi(x)\cup N_{xy}\subset\Pi
(z)$ in ${\cal C}_1$.

For any $w\in\Pi(z)$ in ${\cal C}_1$, if $w\in\Pi(z)$ in ${\cal C}$,
it must also be parent of $x,y$ and $z$ in~${\cal C}_1$, so $w\in\Pi
(x)$ in ${\cal C}_1$. If $w-z$ is an undirected edge in ${\cal C}$,
there exist undirected edges $w-x$ and $w-y$
in ${\cal C}$ such that $w \to z$ is in ${\cal C}_1$. Thus, $w\in
N_{xy}$ in ${\cal C}_1$. We have that $w\in\Pi(x)\cup N_{xy}$ and $\Pi
(z)\subset\Pi(x)\cup N_{xy}$ in ${\cal C}_1$. Thus, $\Pi(z)= \Pi
(x)\cup N_{xy}$ in~${\cal C}_1$, and the condition \textbf{rm}$_2$ holds.

Any undirected path between $x$ and $y$ in ${\cal C}_1$ will also be
an undirected path in ${\cal C}$, so these paths contain at least one
vertex in $N_{xy}$ in ${\cal C}$. From the proof above,
any vertex in $N_{xy}$ in ${\cal C}$ is also a vertex of $N_{xy}$ in
${\cal C}_1$. Thus any undirected path between $x$ and $y$ contains a
vertex in $N_{xy}$ in ${\cal C}_1$, and the condition \textbf{rm}$_3$ holds.
\end{pf*}

\begin{pf*}{Proof of Lemma~\ref{re6}}
From Lemma~\ref{lemma32} and the condition \textbf{rm}$_3$, there
exists a consistent extension of ${\cal C}$,
denoted by $\cal D$, such that all neighbors of $x$ in ${\cal C}$ are
children of $x$ in $\cal D$,
and all neighbors of $y$ in ${\cal C}$ are parents of $x$ in $\cal D$.
Changing $y\to z$ to $z\to y$ in $\cal D$, we obtain a new graph ${\cal
D}'$. From the proof of Lemma~\ref{remov}, we can get that (1) ${\cal
D}'$ is a DAG, (2) ${\cal D}'$ is a consistent extension of~${\cal
C}_1$. Thus, $\cal D$~is a consistent extension of the PDAG that
results from making the $v$-structure $x\to z\leftarrow y$ in ${\cal
C}_1$. Thus, we can get ${\cal C}$ by applying MakeV $x\to z\leftarrow
y$ to ${\cal C}_1$. This implies that MakeV $x\to z\leftarrow y$ is a
valid operator of ${\cal O}_1$ and satisfies the condition mv$_1$.
\end{pf*}

\begin{pf*}{Proof of Theorem~\ref{irr}}
In order to prove this theorem, we first introduce three results:
Lemmas~\ref{delu},~\ref{pppren} and~\ref{rparent}.
\end{pf*}

\begin{lemma}\label{delu}
For any completed PDAG ${\cal C}$ containing at least one undirected
edge, there exists an undirected edge $x- y$ for which $N_{xy}$ is a clique.
\end{lemma}
%
\begin{lemma}\label{pppren}
For any completed PDAG ${\cal C}$, if $x\to y$ occurs in ${\cal C}$,
then $\Pi_x\neq\Pi_y\setminus x$.
\end{lemma}
A proof of Lemmas~\ref{delu} and~\ref{pppren} can be found in
Chickering~\cite{chickering2002learning}.

\begin{lemma}\label{rparent}
For any completed PDAG ${\cal C}$ containing no undirected edges and at
least one directed edge, there exists at least one vertex $x$ for which
any parent of $x$ has no parent.
\end{lemma}
\begin{pf}
The following procedure will find the vertex whose parent has no
parent. Let $a\to b$ be a directed edge in ${\cal C}$, set $y=a$ and $x=b$.
\begin{longlist}[(2)]
\item[(1)] If $\Pi_y$ is not empty, choose any vertex $u$ in $\Pi_y$,
set $x=y$ and $y=u$. Repeat this step until we find a directed edge
$y\to x$ for which $\Pi_y$ is empty.
\item[(2)] Since $\Pi_y$ is empty, from Lemma~\ref{pppren}, there
exists at least one vertex other than $y$ in $\Pi_x$. If there is a
vertex $u\in\Pi_x$ and $u \neq y$ such that $\Pi_u$ is not empty,
choose a vertex in $\Pi_u$, denoted as $v$ and set $y=v$ and $x=u$, and
go to step 1.
\end{longlist}

Since ${\cal C}$ is an acyclic graph with finite vertices, above
procedure must end at the step in which the parents of $x$ have no parents.\vadjust{\goodbreak}
\end{pf}

We now show a proof of Theorem~\ref{irr}.
\begin{pf*}{Proof of Theorem~\ref{irr}}
We need to show that for any two completed PDAGs ${\cal C}_1,{\cal
C}_2\in{\cal S} $, there exists a sequence of operators in ${\cal O}$
such that ${\cal C}_2$ can be obtained by applying a sequence of
operators to PDAGs, starting from~${\cal C}_1$. Because ${\cal O}$ is
reversible, any operator in ${\cal O}$ has a reversible operator, so we
just need to show that any completed PDAG can be transferred to empty
graph without edges. The procedure includes three basic steps.

(1) Deleting all undirected edges.

From Lemma~\ref{delu}, for any completed PDAG containing at least one
undirected edge, we can find an operator with type of DeleteU that
satisfies the condition \textbf{du}$_1$ in Definition \ref
{constrainedset}. We can delete an undirected edge with this operator
and get a new completed PDAG whose skeleton is a subgraph of the
skeleton of the initial completed PDAG. Repeating this procedure, we
can get a completed PDAG, denoted as ${\cal C}_i$, which contains no
undirected edges.

(2) Deleting some directed edges.

From Lemma~\ref{rparent}, we can find a vertex, denoted as $x$, whose
parents have no parents in the completed PDAG ${\cal C}_i$. If $\Pi_x$
contains more than two vertices, we can choose a vertex $u\in\Pi_x$.
Because (1) $N_{x}$ is empty in ${\cal C}_i$, and (2) any other
directed edge $v\to x$ forms a $v$-structure in ${\cal C}_i$, we have
that $v \to x$ is also compelled in the completed PDAG obtained by
deleting directed edge $u\to x$ from
${\cal C}_i$. We can delete $v\to x$ from
${\cal C}_i$ and get a new completed PDAG whose skeleton is a subgraph
of the skeleton of the initial one. Thus, the new completed PDAG is in
$\cal S$. Repeat this procedure for all other directed edges $v'\to x$
in which $v'\in\Pi_x$ until there are only two vertices in $\Pi_x$ in
the new completed PDAG, denoted as ${\cal C}_j$.

(3) Removing a $v$-structure.

The conditions rm$_1$, rm$_2$ and rm$_3$ hold for the $v$-structure
$y\to x\leftarrow u$ in~${\cal C}_j$, so, we can remove $y\to
x\leftarrow u$ from ${\cal C}_j$ and get a new completed PDAG whose
skeleton is a subgraph of the skeleton of the initial graph. Denote the
resulting completed PDAG as ${\cal C}_k$; it may still contain some
undirected edges.

By repeatedly applying the above the steps in sequence, we can finally
obtain a graph without any edges.
\end{pf*}
\end{appendix}

\section*{Acknowledgments}

This work was partly done when Yangbo He was visiting Department of
Statistics in UC Berkeley. Yangbo He would like to thank Prof. Lan Wu
for her support of this visit. Jinzhu Jia's work was done when he was a
postdoc in UC Berkeley. We are very grateful to Adam Bloniarz for his
comments that significantly improved the presentation of our
manuscript. We also thank Jasjeet Sekhon, the co-Editor, the Associate
Editor and the reviewer for their helpful comments and suggestions.

\begin{supplement}
\stitle{Supplement to ``Reversible MCMC on Markov equivalence classes
of sparse directed acyclic graphs''}
\slink[doi]{10.1214/13-AOS1125SUPP} 
\sdatatype{.pdf}
\sfilename{aos1125\_supp.pdf}
\sdescription{In this supplementary note, we give some algorithms,
examples, an experiment and the proofs of the results in this paper.}
\end{supplement}


\printaddresses

\end{document}